%% file: main.tex
\documentclass[conference]{IEEEtran}
\usepackage{times}

\usepackage[noadjust]{cite}
\usepackage{multicol}
\usepackage{booktabs}
\usepackage{multirow}
\usepackage{multicol}
\usepackage{adjustbox}
\usepackage{diagbox}
\usepackage[bookmarks=true]{hyperref}
\usepackage{xcolor}
\usepackage{xspace}
\usepackage{amsmath}
\usepackage{subcaption}
\usepackage{tabularx}
\usepackage{graphicx}

\newcommand{\real}{\texttt{Real}}
\newcommand{\DC}{\texttt{DC}}
\newcommand{\PD}{\texttt{Prior}}
\newcommand{\dc}{task-aware digital cousins}

\begin{document}

\title{Sim-and-Real Co-Training: A Simple Recipe for Vision-Based Robotic Manipulation}

\author{Abhiram Maddukuri$^{*,1}$, Zhenyu Jiang$^{*,1,2}$, Lawrence Yunliang Chen$^{*2,3}$, Soroush Nasiriany$^{*1,2}$,\\Yuqi Xie$^2$, Yu Fang$^2$, Wenqi Huang$^2$, Zu Wang$^{2,4}$, Zhenjia Xu$^2$, Nikita Chernyadev$^2$,\\Scott Reed$^2$, Ken Goldberg$^3$, Ajay Mandlekar$^{\dagger,2}$, Linxi Fan$^{\dagger,2}$, Yuke Zhu$^{\dagger,1,2}$\vspace{-2mm}\\\\$^1$UT Austin ~ $^2$NVIDIA ~ $^3$UC Berkeley ~ $^4$New York University\vspace{-3mm}}

\maketitle

\begin{abstract}

Large real-world robot datasets hold great potential to train generalist robot models, but scaling real-world human data collection is time-consuming and resource-intensive. Simulation has great potential in supplementing large-scale data, especially with recent advances in generative AI and automated data generation tools that enable scalable creation of robot behavior datasets.
However, training a policy solely in simulation and transferring it to the real world often demands substantial human effort to bridge the reality gap. A compelling alternative is to \textit{co-train} the policy on a mixture of simulation and real-world datasets. 
Preliminary studies have recently shown this strategy to substantially improve the performance of a policy over one trained on a limited amount of real-world data. Nonetheless, the community lacks a systematic understanding of sim-and-real co-training and what it takes to reap the benefits of simulation data for real-robot learning. 
This work presents a simple yet effective recipe for utilizing simulation data to solve vision-based robotic manipulation tasks.
We derive this recipe from comprehensive experiments that validate the co-training strategy on various simulation and real-world datasets. Using two domains---a robot arm and a humanoid---across diverse tasks, we demonstrate that simulation data can enhance real-world task performance by an average of $38\%$, even with notable differences between the simulation and real-world data. 
Videos and additional results can be found at \textcolor{teal}{\href{https://co-training.github.io/}{co-training.github.io}}.

\end{abstract}

\IEEEpeerreviewmaketitle

\newcommand\blfootnote[1]{%
  \begin{NoHyper}%
  \renewcommand\thefootnote{}\footnote{#1}%
  \addtocounter{footnote}{-1}%
  \end{NoHyper}%
}

\blfootnote{$^*$Equal contribution.  $^\dagger$Project leads.}

\input{sections/00_intro}

\input{sections/01_related_work}

\input{sections/02_problem}

\input{sections/03_study}

\input{sections/04_experiments}

\input{sections/06_limitations}

\input{sections/05_conclusion}

\clearpage
\bibliographystyle{IEEEtran}
\bibliography{IEEEabrv,references}

\clearpage
\input{sections/appendix}

\end{document}

%% file: sections/00_intro.tex
\section{Introduction}
\label{sec:introduction}

\begin{figure}[t]
    \centering
    \includegraphics[width=\linewidth]
    {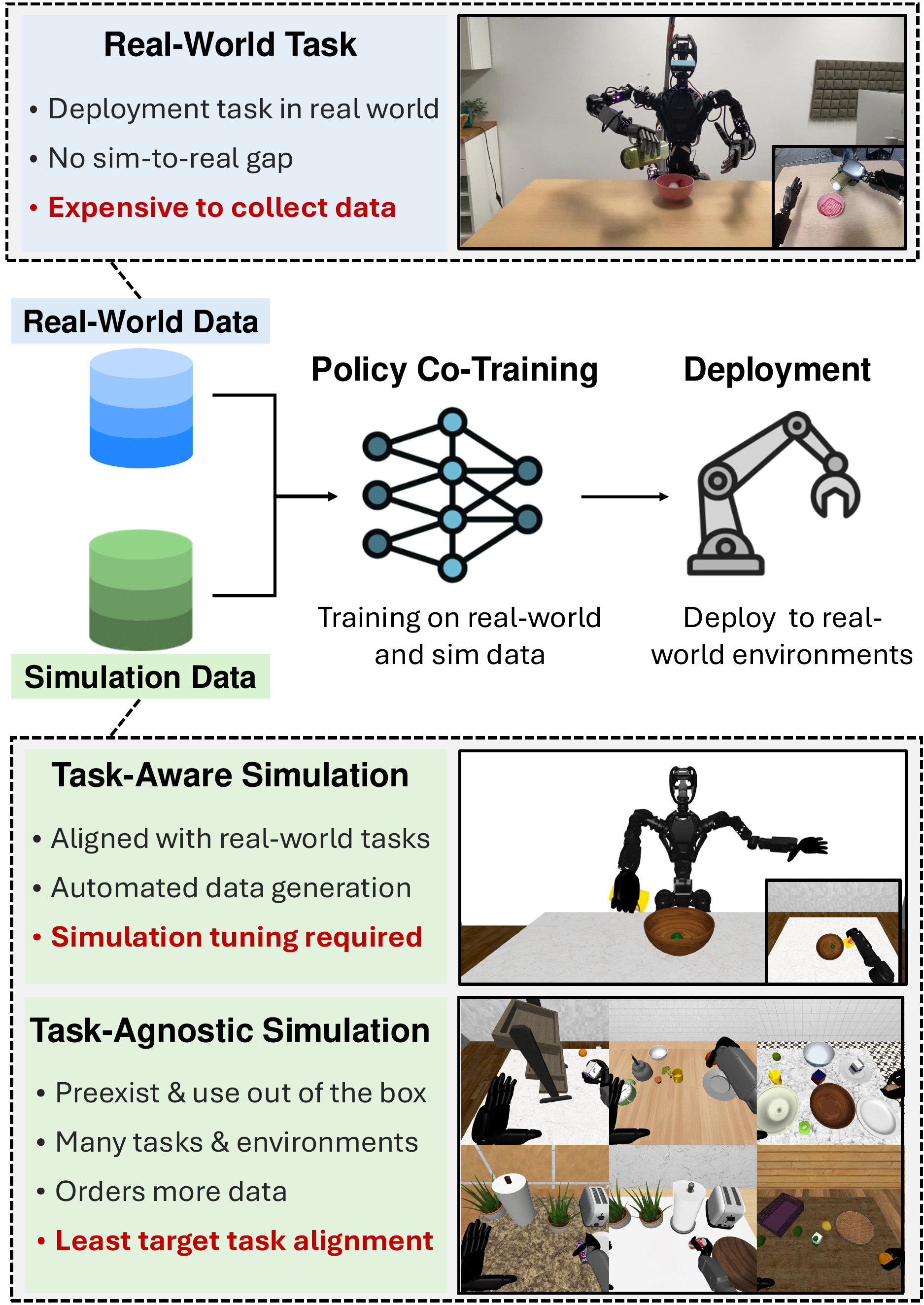}
    \caption{\small{\textbf{Sim-and-Real Co-Training.} We show how co-training policies on real-world and simulation data can attain superior performance in the real-robot deployment, compared to training solely on real-world data. We specifically study two forms of simulation data: (1) task-aware data from \textit{digital cousins} built with knowledge of the real-world tasks, and (2) task-agnostic data from multi-task \textit{prior simulations} covering more diverse settings but with less alignment to the target task.}}
    \label{fig:pull}
    \vspace{-10pt}
\end{figure}

The ability to generalize across diverse environments and tasks is a critical step toward realizing generalist robotic systems. 
Recent advancements in robot foundation models~\cite{black2024pi_0,kim2024openvla,bjorck2025gr00t}---trained on a mixture of web-scale vision-language datasets and robot-specific datasets---have demonstrated significant potential for cross-domain generalization.
Large real-world robot datasets~\cite{ebert2021bridge, brohan2022rt} embody this diversity and are a crucial source of data.
Despite this progress, challenges remain in achieving reliable real-world deployment.
To bridge this gap, there have recently been multiple efforts on collecting even larger-scale real-robot datasets~\cite{khazatsky2024droid, black2024pi_0}. These works have shown the potential of data-driven methods in acquiring versatile robotic skills. However, they involve considerable cost, time, and scalability challenges, and it remains unclear whether simply scaling real-world data collection alone is sufficient to train generalist robot models.

Simulation is a promising alternative to mitigate the data hunger of large models. 
The recent proliferation of generative AI tools allows for the automated generation of assets, scenes, and tasks in simulation, all of which can be produced with high-fidelity physics simulators and photorealistic renderers~\cite{robocasa2024, wang2023robogen}. 
Furthermore, automated data generation tools can be applied to these simulation environments to synthesize large amounts of diverse, high-quality robot trajectories with minimal human effort~\cite{mandlekar2023mimicgen, jiang2024dexmimicen, garrett2024skillmimicgen, dalal2023imitating}, offering massive training data for generalist manipulation policies. 
However, approaches that use simulation data must deal with the reality gap since the visuals and physics in simulation do not align perfectly with the real world. 
Prior approaches on sim-to-real policy transfer typically rely on extensive tuning of simulation to match the real world~\cite{ramos2019bayessim,muratore2022neural,lim2022real2sim2real,memmel2024asid,ren2023adaptsim}, or meticulously randomizing a specific set of simulation parameters~\cite{tobin2017domain,peng2018sim,zhu2018reinforcement,andrychowicz2020learning,handa2022dextreme}. 
Such approaches can require significant human effort.

A compelling alternative to sim-to-real transfer is to directly co-train policies on a \textit{mixture} of simulation and real-world data. 
Preliminary findings in recent work~\cite{bousmalis2018using, robocasa2024, ankile2024imitation} suggest that incorporating simulation data in this way can greatly improve policy performance compared to using real-world data alone. 
Moreover, sim-and-real co-training may not require the high level of alignment between simulation and reality typically needed for sim-to-real transfer, making it a promising strategy to tap into the potential of large synthetic datasets with minimal human effort. 
Despite its promise, the community lacks a systematic understanding of this strategy and what it takes to reap the benefits of simulation data for real-robot learning. It remains unclear how different the simulation data can be from the real-world data, and what kinds of dataset mixtures and compositions are ideal.

\textbf{This work presents a simple recipe for supplementing real robot datasets with synthetic simulation datasets to facilitate learning vision-based manipulation policies for real robots.} 
We derive this recipe by conducting a comprehensive set of experiments that co-train robot policies on various simulation and real-world datasets.
As shown in Figure~\ref{fig:pull}, we focus on two concrete sources of simulation data --- task-aware simulation, in which the simulation environment is intentionally designed to align with the real world loosely, akin to the ``digital cousin'' concept introduced in Dai et al.~\cite{dai2024automated}, and task-agnostic simulation, which comprises prior simulation data made independently of the particular target task.
Our study is carried out on two distinct robot embodiments (robot arm and humanoid) across several diverse tasks, spanning pick-and-place, articulated object manipulation, and non-prehensile manipulation (\textit{e.g.}, pouring). We investigate a number of critical data composition factors to understand the degree to which simulation and real-world data must be aligned for co-training to be effective. For example, should the tasks, scenes, and objects be the same between the sim and the real? How about the location of workspace cameras and object placements? We make use of synthetic data generation tools~\cite{mandlekar2023mimicgen, jiang2024dexmimicen} to test different simulation dataset compositions with ease, resulting in actionable insights for robotics practitioners.

\textbf{We summarize our contributions as follows:}
\begin{enumerate}
    \item We establish a systematic study for co-training on real-robot data and synthetically generated data from simulation, resulting in a simple recipe to leverage synthetic simulation data for real-world manipulation;
    \item We demonstrate empirically how co-training on synthetic simulation data can be broadly useful in facilitating policy learning for downstream real-world tasks, improving policy performance across two domains by an average of $38\%$;
    \item We derive insight into what types of simulation data are most effective for sim-and-real co-training. Surprisingly, we find that simulation data provides substantial benefits even with notable differences from the real-world data, and that diverse simulation data can facilitate generalization to unseen scenarios in the real world.
\end{enumerate}

%% file: sections/01_related_work.tex
\section{Related Work}
\label{sec:related_work}

\subsection{Learning Manipulation from Demonstration Data} 

Behavior cloning~\cite{pomerleau1989alvinn} is a widely adopted approach for learning robot policies from demonstration data~\cite{drolet2024, tung2020learning, zhao2023learning, aldaco2024aloha,schaal1999imitation, Ijspeert2002MovementIW, seo2023deep, ding2024bunny, cheng2024open,robomimic2021, chi2023diffusion}. In this framework, policies are trained to predict actions based on ground truth state-action pairs provided in a demonstration dataset. This method has been extensively applied in robot manipulation tasks~\cite{finn2017one, Billard2008RobotPB, Calinon2010LearningAR, mandlekar2020learning, zeng2020transporter, wang2021generalization, lynch2019learning,pertsch2021skild,ajay2020opal,hakhamaneshi2021fist,zhu2022bottom,nasiriany2022sailor}. However, its success in real-world applications typically hinges on the availability of large amounts of high-quality demonstration data, which can be prohibitively expensive to collect. To address this limitation, our work explores the use of simulation data to enhance real-world robot manipulation through imitation learning, thereby reducing the dependency on costly real-world data collection.

\subsection{Sim-to-Real and Sim-Real Co-Training}
Sim-to-real transfer has been a pivotal focus in robotics research, aimed at enabling models trained in simulation to perform effectively in the real world. 
One popular approach is domain randomization~\cite{tobin2017domain,peng2018sim,zhu2018reinforcement,andrychowicz2020learning,chebotar2019closing,mehta2020active,handa2022dextreme}, which introduces variability into the simulation environment to train policies that are robust to discrepancies between simulation and reality. 
However, domain randomization approaches can require careful tuning and a significant human burden to determine proper randomization ranges for the parameters that enable the policy to transfer to the real world.
Another common approach seeks to minimize the sim-to-real gap by improving simulation fidelity to match the real world closely. Techniques such as system identification~\cite{ramos2019bayessim,huang2023went,muratore2022neural,lim2022real2sim2real,kumar2021rma,memmel2024asid,ren2023adaptsim,evans2022context} and the creation of digital twins~\cite{jiang2022ditto,torne2024reconciling} aim to align simulated dynamics more closely with real-world conditions. These methods often require significant human effort, preventing their applicability to diverse tasks and environments.
Instead, recent research has trained real-world manipulation policies using a mixture of simulation and real-world data~\cite{bousmalis2018using, ankile2024imitation,robocasa2024,brohan2023rt,bjorck2025gr00t}, and demonstrated superior performance to solely using the same quantity of real-world data. Furthermore, the simulation data in these approaches does not necessarily need to be perfectly aligned with the real world, making it a compelling alternative to other approaches.
Building on these findings, our work systematically investigates sim-and-real co-training. We examine the effectiveness of co-training real-world policies using two sources of simulation data with varying levels of alignment: task-aware digital cousins~\cite{dai2024automated} and task-agnostic prior simulation, as shown in Figure~\ref{fig:pull}. Our study offers practical insights and methodologies for practitioners aiming to achieve robust performance in real-world scenarios.

\subsection{Dataset Composition in Robot Learning} 
Recent research~\cite{hejna2024re,xie2024decomposing} has highlighted the importance of dataset composition in robot learning, particularly in understanding how variations in data quality and diversity influence policy generalization~\cite{pumacay2024colosseum,gao2024efficient}. Studies such as MimicLabs~\cite{saxena2024matters} have conducted large-scale analyses to identify the types of data that maximize the utility of robotic datasets and improve downstream policy performance. Inspired by this line of work, we investigate the optimal composition of simulation and real-world data specifically for real-world robotic manipulation tasks. Our study aims to provide actionable guidelines on how to strategically combine these data sources to achieve superior policy learning outcomes in the real world.

%% file: sections/02_problem.tex
\begin{figure*}[t]
    \centering
    \includegraphics[width=1.0\linewidth]
    {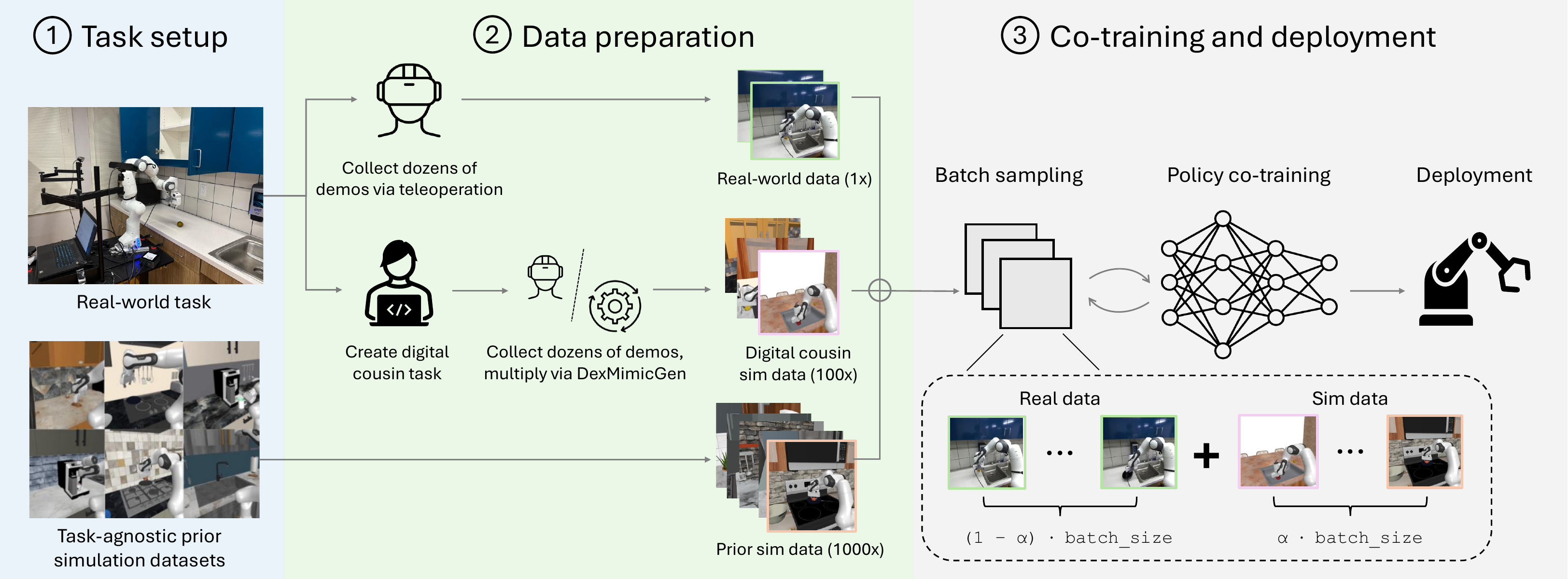}
    \caption{\small{\textbf{Method Overview.} Our workflow consists of three components: (1) We start with a real-world target task in mind and some prior simulation data; (2) Given real-world tasks and prior simulation data, we build simulated digital cousin environments that share semantic similarities with their real-world counterparts but may still hold discrepancies in visual and physical aspects. We leverage synthetic data generation methods to multiply trajectories in digital cousins, producing a large quantity of demonstrations in simulation. From here, we consolidate prior simulation data, digital cousin data, and real-world data; (3) We co-train the policy on a mixture of real-world and simulation data. We sample simulation data according to a sampling ratio of $\alpha$, which is crucial for the method's effectiveness. After training the policy, we deploy the learned policy directly to the real robot.}}
    \label{fig:setup}
\end{figure*}

\section{Problem Statement and Preliminaries}
\label{sec:problem}

\subsection{Co-Training on Real-World and Simulation Data} \label{subsec:cotraining_ratio_prelim}

We assume access to robot trajectory demonstrations collected in real environments, $\mathcal{D}_{\text{real}} = \{ \xi_i \}_{i=1}^N$.
Instead of training a policy on demonstrations solely from the real world, we have additional demonstrations from simulation environments, $\mathcal{D}_{\text{sim}} = \{ \xi_i \}_{i=1}^M$, where generally $M \gg N$.
We train a visuomotor policy $\pi_{\theta}$ on these two data sources.
We adopt the \textit{co-training} formulation following prior work \cite{robocasa2024}, where we minimize the behavioral cloning action loss
\begin{equation}
\label{eq:loss}
    \mathcal{L}_{\text{total}}(\theta; \mathcal{D}_{\text{real}}, \mathcal{D}_{\text{sim}}) = \alpha \cdot \mathcal{L}(\theta; \mathcal{D}_{\text{sim}}) + (1 - \alpha) \cdot \mathcal{L}(\theta; \mathcal{D}_{\text{real}})
\end{equation}
where $\mathcal{L}(\theta; \mathcal{D}) = \frac{1}{|\mathcal{D}|} \sum_{(o_i, a_i) \in \mathcal{D}} -\log \pi_\theta(a_i | o_i)$ and $\alpha \in [0, 1]$ is the \textit{co-training ratio} balancing the relative weight of simulation and real-world data. 
In practice, we use an equivalent formulation of $\alpha$, which represents the probability of sampling from simulation data in each training batch. Specifically, we reweight each sample $(o_i, a_i)$ such that the probability of drawing it from the simulation dataset is $P[(o_i, a_i) \in \mathcal{D}_{\text{sim}}] = \alpha$, while the probability of drawing it from the real dataset is $P[(o_i, a_i) \in \mathcal{D}_{\text{real}}] = 1 - \alpha$ during training batch sampling.
We further detail the relative weighting procedure in Appendix~\ref{app:training}.
As we will see in experiments, the choice of $\alpha$ is crucial to policy performance.
Our end objective is to produce vision-based manipulation policies that maximize task performance on one or multiple downstream tasks in real-world environments.

\subsection{Data Composition Factors}
$\mathcal{D}_{\text{real}}$ and $\mathcal{D}_{\text{sim}}$ can comprise demonstration trajectories from either a single task or a diverse array of tasks, embodiments, and environments.
To reason about how the particular choices in constructing these datasets can affect the success of co-training, it is useful to decompose these datasets into a set of \textit{data composition factors}.
We assume that each dataset follows a distribution of factors $\{\mathcal{Z}^{(1)}, \mathcal{Z}^{(2)}, \cdots, \mathcal{Z}^{(K)}\}$, borrowing notation from recent work~\cite{saxena2024matters}.
We do not assume that the simulation dataset is perfectly aligned with the real-world dataset, \textit{i.e.}, for some factors $\mathcal{Z}^{(i)}_{\text{sim}} \neq \mathcal{Z}^{(i)}_{\text{real}}$.
Despite these alignment gaps, we are interested in transferring knowledge from simulation domains to learn a more effective policy $\pi$ for real-world tasks.

Common data composition factors include, but are not limited to, the following:

\begin{itemize}
    \item \noindent\textbf{Task composition:} Which tasks, and by extension, subtasks and motions, are present in the simulation and real-world data. Even if the real-world and simulation data involve solving the same task, there may be multiple valid ways to achieve the task. Different datasets may exhibit different orderings of subtasks and different manipulation skills;

    \item \noindent\textbf{Scene composition:} The number of scenes in simulation and real-world data, in addition to the scope and diversity of various components across these scenes. For example, the number of fixtures, articulation properties of interactive objects, range of lighting conditions, and range of background textures;

    \item \noindent\textbf{Object composition:} Which object categories are present in the simulation and real-world data, and the number of unique object instances per object category;

    \item \noindent\textbf{Initialization distribution:} The initial state distribution in the datasets, representing the distribution of states in the initial state of each trajectory in the dataset. This comprises the initial robot base pose and arm joints, in addition to the initialization distribution of objects and fixtures in the scene;

    \item \noindent\textbf{Camera parameters:} We assume that we have a set of $N$ cameras used to train each visuomotor agent. Each camera has a range of values across several parameters. The most prominent parameters are camera intrinsics and camera extrinsics;

    \item \noindent\textbf{Dynamics parameters:} Key physical parameters such as friction, mass of objects, and inertia. Other parameters include robot controller variables such as the type of controller and its gains.
\end{itemize}

We define these parameters in more detail and quantify them in Section~\ref{sec:study_setup}, when we introduce the domains and tasks, and we study how important it is to align each factor between simulation and the real world for co-training success.

\subsection{Automated Synthetic Data Generation}
A key advantage of simulation is ease of data collection---we leverage automated synthetic data generation tools to generate large, high-quality simulation datasets and use them for co-training with smaller real-world datasets.
For each task in the simulation, we first collect dozens of source human demonstrations.
We then use MimicGen~\cite{mandlekar2023mimicgen} to generate large synthetic trajectory datasets at scale.
For bimanual and humanoid robots, we use DexMimicGen~\cite{jiang2024dexmimicen}, a method that builds on top of MimicGen.
The process is as follows. 
First, we segment these source demonstrations into a sequence of object-centric segments.
(Dex)MimicGen then generates new demonstrations by applying linear transformations to selected source demonstration segments and concatenating these transformed segments to form novel trajectories. By leveraging these methods, we can use physics simulations to multiply the number of trajectories by orders of magnitude.

%% file: sections/03_study.tex
\begin{figure*}
    \centering
    \includegraphics[width=1.0\linewidth]{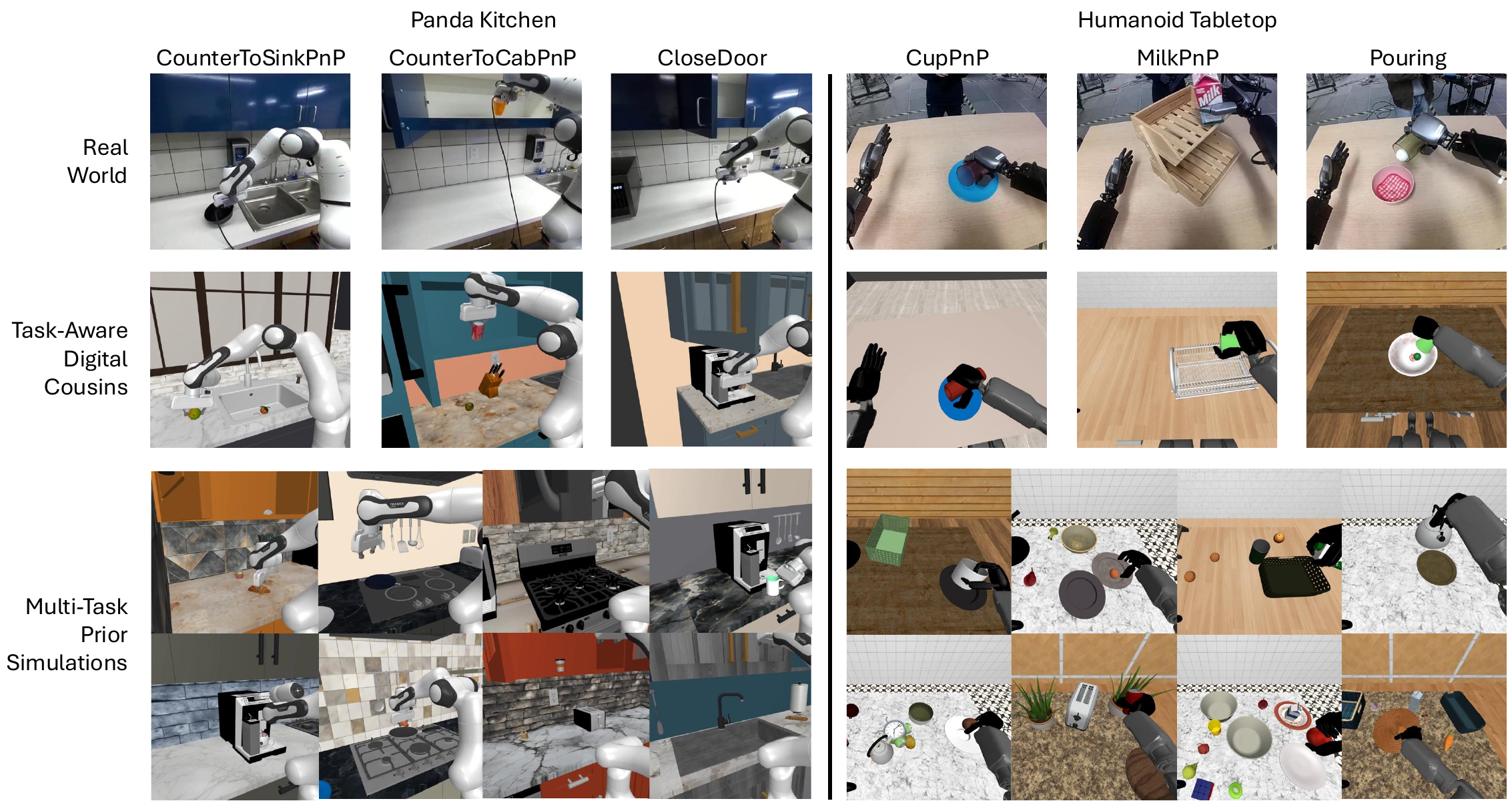}
    \caption{\small{\textbf{Real-World and Simulation Tasks.} We experiment with co-training on three data sources on the two robot domains of Kitchen Panda and Humanoid Tabletop: (top) data collected for real-world tasks; (middle) data from task-aware digital cousin environments that resemble the target tasks but are not perfectly aligned; (bottom) prior multi-task data from simulation that comprise a wide range of tasks and environments but have larger discrepancies with real-world tasks.}}
    \label{fig:task}
\end{figure*}

\section{Study Setup}
\label{sec:study_setup}
Our goal is to develop a simple recipe for co-training on real-robot and simulation data to significantly improve real-world policy performance compared to training on real data alone.
We are broadly interested in two scenarios:
\\ \\
\noindent (1) \textbf{Co-training with prior large-scale simulation data.}
Can we use existing large prior simulation datasets as co-training data?
Note, these datasets often have significant discrepancies with the real world in terms of visual features, task semantics, and behaviors.
We are interested in understanding the extent to which these datasets can help \textit{out of the box} in learning downstream real-world tasks in spite of these domain gaps.
\\ \\
\noindent (2) \textbf{Co-training with task-aware simulation data.} Given knowledge of the real-world tasks, we can create customized simulation datasets that are better aligned with the real-world tasks.
However, tuning simulation environments to match the real-world environments precisely is impractical.
Which data composition factors are most important to align between simulation and the real-world setup, and can we forgo \textit{perfect alignment}, allowing us to reduce human effort?
\\ \\
See Figure~\ref{fig:setup} for an overview of our workflow, including the real-world setup, simulation pipeline, and the co-training procedure. We describe all of these components in detail in the following sections.

\subsection{Real-World Domains}
\label{sec:study-real}
We seek a co-training recipe that is broadly applicable to a wide range of embodiments, tasks, and environments.
To this end, we conduct a comprehensive study featuring two distinct domains, each with a unique robot embodiment and diverse tasks (see Figure~\ref{fig:task} for an illustration):
\\ \\
\noindent\textbf{Panda Kitchen.}
A real-world kitchen environment with the Franka Emika Panda robot. We adopt the DROID tabletop hardware setup~\cite{khazatsky2024droid}, with some minor modifications (see Appendix~\ref{app:setup} for details).
We experiment with three real-world tasks and collect 50 human demonstrations for each task: 
\begin{itemize}
    \item \texttt{CounterToSinkPnP}: move an object from the counter to the sink basin. This task features nine object categories with diverse shapes: can, cup, coffee cup, water bottle, lemon, garlic, bowl, granola bar, and pear.
    \item \texttt{CounterToCabPnP}: move an object from the counter to the cabinet. This task features eight object categories.\footnote{We use the same object categories as CounterToSinkPnP but exclude the water bottle due to difficulties in placing it stably into the cabinet.}
    \item \texttt{CloseDoor}: close the door for an overhead cabinet.
\end{itemize}

\noindent\textbf{Humanoid Tabletop.} A real-world tabletop environment with a Fourier GR-1 humanoid robot. We control the robot using a mink-based~\cite{Zakka_Mink_Python_inverse_2024} IK controller. We use a first-person view RGB camera mounted to the head of the humanoid. We choose three tasks and collect 20 human demonstrations for each task. We describe more details in Appendix~\ref{app:setup}.
\begin{itemize}
    \item \texttt{CupPnP}: move a cup from the plate to the table.
    \item \texttt{MilkPnP}: move a box of milk from the table to the second level of the shelf.
    \item \texttt{Pouring}: pick up a cup with a ping-pong ball inside and pour the ball into a bowl on the table.
\end{itemize}

Our study is grounded in the real world---we compare the efficacy of different co-training methods by directly evaluating policies on these real-world tasks.

\subsection{Prior Task-Agnostic Simulation Data}
\label{sec:study-prior}

We leverage synthetically generated data to supplement real-robot datasets for policy training.
One approach is to directly co-train with existing large-scale simulation datasets, or \textit{prior task-agnostic simulation datasets}. We define a \textit{prior task-agnostic simulation dataset} as any simulation dataset that existed before the creation of the downstream, real-world task. For the purpose of direct co-training, we use prior task-agnostic datasets that contain the same robot embodiment and action space, but this is not a strict requirement. We otherwise assume that these datasets cover a broad range of tasks and environments.
We are interested in co-training with these datasets \textit{out of the box}, without expending additional efforts designing new tasks in simulation and collecting new data.
These datasets may have numerous discrepancies with real-world data, but they present a simple and convenient way to leverage simulation data.
We use the following prior simulation datasets:\\

\noindent\textbf{Panda Kitchen.} We use the multi-task RoboCasa dataset~\cite{robocasa2024}.
We choose RoboCasa, given its focus on kitchen environments, a diverse range of scenes and tasks, and the availability of large robot data.
In addition, Nasiriany et al.~\cite{robocasa2024} showed preliminary findings that co-training with simulation data can aid transfer in a real-world kitchen environment.
The dataset comprises 72k demonstrations across 24 tasks and 100 scenes; for each task, 3,000 demonstrations were generated from 50 source human demonstrations using the MimicGen data generation system~\cite{mandlekar2023mimicgen}.
Refer to Appendix~\ref{app:priordata} for in-depth details about the tasks and datasets.
Note that three of these tasks semantically correspond to our real-world tasks but include notable discrepancies with the real-world setup, including initial robot joint positions, controller parameters, physical parameters, object categories, and the robot base position.
We present a quantitative comparison of data composition for the real-world and prior simulation datasets in Appendix~\ref{app:datacomposition}.

Camera alignment differences between simulation and real-world data can be a major discrepancy.
We address this discrepancy by re-rendering the simulation demonstrations to approximately match the camera poses of the real-world setup.\footnote{This operation introduces occlusions for the drawer and stove knob manipulation tasks, so we opt to exclude these. In total, we use 60k demonstrations across 20 tasks.}
Note that this does not represent perfect alignment, but, as we demonstrate in our experiments, it can significantly help nonetheless.
Beyond this simple post-processing operation, we do not make any further changes to the prior data.\\
\\
\noindent\textbf{Humanoid Tabletop.}
To mirror the setup for the kitchen domain, we create a prior task-agnostic dataset comprising 10 tasks in RoboCasa involving a single kitchen countertop and a GR-1 robot.
Each task involves grasping a specified object from a source receptacle and placing it into a target receptacle (\textit{e.g.}, from bowl to basket).
Refer to Appendix~\ref{app:priordata} for additional details about these tasks.
While semantically similar to the real-world setup, 
the prior tasks and datasets were developed independently and involve numerous discrepancies such as object categories, visual textures, distractor objects, and physical parameters.
None of these 10 tasks is semantically equivalent to the real-world tasks---they involve different source and/or target receptacles.
We use DexMimicGen~\cite{jiang2024dexmimicen}, a data generation framework built on top of MimicGen for humanoid and other bimanual robots, to synthesize robot trajectories from dozens of human demonstrations.
We generate 1,000 demonstrations for each task, resulting in 10k total demonstrations.

\subsection{Building Task-Aware Simulation Datasets}
\label{sec:study-dc}
The tasks and datasets presented in the previous section may have a number of large discrepancies with the real-world tasks, potentially limiting their utility.
Alternatively, we can expend additional effort in creating custom tasks in simulation that are better aligned with the real-world tasks.
Creating a perfect \textit{digital twin}~\cite{nasiriany2022maple,jiang2022ditto,torne2024reconciling} copy of the real-world task is challenging, requiring extensive manual tuning, system identification, and sourcing identical 3D assets.
Instead, we opt to create tasks in simulation that share the same task semantics, namely the object categories in the environment and the same behaviors.
We refer to these as \textit{task-aware digital cousins}.
The term ``digital cousin'' was recently introduced by Dai et al.~\cite{dai2024automated} to describe simulation environments that are close to, but not perfectly aligned with, their real-world counterparts. We extend this notion with a more precise definition: a \textit{task-aware digital cousin} is a simulation dataset that preserves four key elements of the real-world task:
\begin{enumerate}
    \item The same robot and action space;
    \item The same task goal—specifically, the same success check and, if applicable, the same language instructions;
    \item The same object categories, though individual instances may differ in geometry or texture;
    \item The same environmental fixture categories (\textit{e.g.}, kitchen counters, tabletops, cabinet doors).
\end{enumerate}
We outline our efforts to create these tasks as follows:
\\ \\
\noindent\textbf{Panda Kitchen.} The real-world Panda Kitchen tasks are already represented in the RoboCasa prior dataset, but have several discrepancies as outlined in the prior section.
We outline the changes that we made to the task and dataset as follows.
First, we adjust the initial state distribution of the robot joints and robot base position in simulation to match the real environment.
In addition, we restrict the objects in the task to a curated list of 10 object categories, which includes all of the nine object categories used in the real-world \texttt{CounterToSinkPnP} task.
This is in contrast to the prior dataset, in which the authors feature a wider range of 66 possible object categories.
For each task, we then collect 100 source human demonstrations.
Finally, we generate 10,000 demonstrations for each task using MimicGen.
This is in contrast to the prior dataset, where the authors collected 50 source human demonstrations per task and generated 3,000 demonstrations per task with MimicGen.
We provide a more in-depth comparison between the real data, task-agnostic prior simulation data, and task-aware digital cousin simulation data in Appendix~\ref{app:datacomposition}.
\\ \\
\noindent\textbf{Humanoid Tabletop.} For each of the three tasks in this domain, we construct a digital cousin of the real-world environment in RoboCasa. In each of the real-world tasks, we use a fixed set of objects for both data collection and evaluation. In the digital cousin, however, we randomly select objects from the same category as those in the real-world task to increase the diversity of simulation demonstrations. Additionally, we align the robot's initial pose and camera position to closely replicate the real-world setup. We then collect 10 source demonstrations and generate 1,000 trajectories for each task using DexMimicGen~\cite{jiang2024dexmimicen}. A detailed analysis of the data composition is provided in Appendix~\ref{app:datacomposition}.

\begin{table*}[h]
\centering
\small
\setlength{\tabcolsep}{4pt}
\begin{tabular}{l|ccc|ccc|c}
\toprule
\multirow{2}{*}{Data Composition} & \multicolumn{3}{c}{Panda Arm} & \multicolumn{3}{c|}{GR-1 Humanoid} & \multirow{2}{*}{\,Average\,} \\
    \cmidrule(lr){2-4} \cmidrule(lr){5-7}
     & \texttt{C2SPnP} & \texttt{C2CPnP} & \texttt{CloseDoor}  & \texttt{CupPnP} & \texttt{MilkPnP} & \texttt{Pouring}  &  \\
    \midrule
    \texttt{Real} &  44\% & 38\% & 10\% & 65\% & 50\% & 65\%  & 45.3\%\\
    \texttt{Real} + \texttt{DC} & 67\% & \textbf{72\%} & \textbf{100\%}  & \textbf{95\%} & 70\%  & 85\% & 81.1\% \\
    \texttt{Real} + \texttt{Prior}  & 58\% & 53\% & \textbf{100\%} & 80\% & \textbf{80\%}  & 70\% & 76.8\% \\
    \texttt{Real} + \texttt{DC} + \texttt{Prior}  & \textbf{72\%}  & \textbf{72\%} & \textbf{100\%} & 85\%  &  \textbf{80\%} & \textbf{90\%} & \textbf{83.2\%} \\
    \bottomrule 
\end{tabular}
\vspace{5pt}
\caption{\small{\textbf{Effect of different simulation data in the co-training mix.} We compare co-training with different simulation data on six tasks across two robot platforms. Note that we abbreviate the \texttt{CounterToSinkPnP} task as \texttt{C2SPnP} and \texttt{CounterToCabPnP} as \texttt{C2CPnP}. Co-training with \PD{} (third row) consistently boosts performance over policies trained on real data only (first row). On top of these results, adding better aligned \DC{} further improves the overall performance (last row).
}}
\label{tab:main_result}
\end{table*}

\subsection{Training and Evaluation Protocol}

In our study, we compare the effect of co-training with different forms of real-world and simulation data. For each task, we have access to the following forms of data:

\begin{itemize}
    \item \textbf{Real-world data (\texttt{Real})}: demonstrations collected for the target task in the real-world. See Section~\ref{sec:study-real} for the list of tasks.
    \item \textbf{Prior simulation data (\texttt{Prior})}: task-agnostic simulation data outlined in Section~\ref{sec:study-prior}.
    \item \textbf{Task-aware digital cousin data (\texttt{DC})}: synthetic simulation data outlined in Section~\ref{sec:study-dc}.
\end{itemize}

See Appendix Tables~\ref{tab:panda-kitchen-datasets} and ~\ref{tab:humanoid-tabletop-datasets} for an overview and comparison of these datasets for the Panda Kitchen and Humanoid Tabletop domains.

We compare various mixtures of these datasets by co-training a policy on the data and evaluating the resulting policy on our real-world tasks outlined in Section~\ref{sec:study-real}.
We train visuomotor policies with the Diffusion Policy implementation from Chi et al.~\cite{chi2023diffusion}.
The policy takes RGB images and robot proprioceptive information as input and produces a sequence of actions to execute.
We outline specific settings and hyperparameters in detail in Appendix~\ref{app:training}.
Following training, we evaluate the policy across a number of trials and record the success rate.
See Appendix~\ref{app:evaluation} for details on our evaluation protocol.

%% file: sections/04_experiments.tex
\section{Experiments}
\label{sec:experiments}

In this section, we present a comprehensive empirical study of co-training real-world policies using simulation data. We begin by showcasing the benefits of co-training using our full-fledged recipe, which has been informed by systematic experimentation (Section~\ref{subsec:effectiveness} and Section~\ref{subsec:generalization}). Specifically, we demonstrate how co-training with simulation data enhances the real-world policy's in-domain performance (Section~\ref{subsec:effectiveness}) and improves its generalization to novel scenarios (Section~\ref{subsec:generalization}). 

Next, we delve into the systematic experiments that guided the development of our recipe (Section~\ref{subsec:cotrain_factors}). These experiments identify key elements for effective sim-and-real co-training, providing insights into what factors matter most. Finally, we conclude with a concise and actionable recipe based on our findings (Section~\ref{subsec:recipe}), ensuring clarity for practitioners.

\subsection{Effectiveness of Sim-and-Real Co-Training}
\label{subsec:effectiveness}

\textbf{Co-training with task-aware digital cousin data significantly enhances real-world performance beyond real-only policies.} Table~\ref{tab:main_result} presents our main results. In the second row, compared to policies trained only on \real{}, those trained on \real{} and \DC{} exhibit a 35.8\% higher average success rate. These results indicate that incorporating simulation data more closely aligned with real-world tasks significantly enhances real-world policy performance. It is important to note that \DC{} data is generated in task-aware digital cousin simulation environments. These environments share the same task definitions, similar scene setups, and comparable camera views with the real world, though none are perfectly aligned. Nevertheless, with minimal effort, we were able to construct and approximately align these digital cousins with real-world environments (see Section~\ref{sec:study-dc}), demonstrating the feasibility of leveraging such simulations for improved real-world policy training.

\textbf{Co-training with task-agnostic prior simulation data also improves real-world performance.} As shown in the third row of Table~\ref{tab:main_result}, policies trained on \real{} and \PD{} consistently outperform those trained solely on \real{} across all tasks, achieving an average success rate improvement of 31.5\%. This is a particularly surprising and encouraging result, as the \PD{} data are generated without any prior knowledge of real-world tasks. These results indicate that even without manual alignment of the simulation environment, co-training with simulation data yields substantial benefits. This finding highlights the potential of leveraging readily available simulation data to enhance real-world policy performance.

Finally, in the last row of Table~\ref{tab:main_result}, policies trained on the combination of \real{}, \DC{}, and \PD{} data in general perform the best, achieving an improvement of 37.9\% over the real-only policies on average. 

We observe a dramatic performance gap between \real{} and the other co-trained polices for the \texttt{CloseDoor} task. We further investigate the robustness of this gap by training the \real{} policy with more demos. See Section ~\ref{app:closedoorextrademos} for more results.

\subsection{Generalization Beyond Real Demonstrations}
\label{subsec:generalization}

To understand how simulation data enhances real-world policy performance, we investigate whether exposure to diverse situations in simulation—ones not explicitly covered in real-world demonstrations—can improve a policy’s ability to generalize to similar, unseen situations in the real world. This question is particularly important because generating broad-coverage data in simulation is relatively easy, whereas collecting diverse real-world demonstrations is often expensive and impractical. If simulation data can effectively bridge this gap, it would provide a powerful and scalable way to improve real-world policies with minimal real-world data.

To explore this, we evaluate the generalization capabilities of co-trained policies beyond real-world demonstrations. Specifically, we consider two key axes of variation: novel objects and novel initial positions. We use the \texttt{CounterToSinkPnP} and \texttt{CupPnP} tasks as our testbed, where, by default, these factors are randomized in simulation to assess the policy’s ability to handle novel scenarios in the real world (see Section~\ref{sec:study-dc}). %

\textbf{Co-training with simulation data enhances policy robustness to novel object entities.} For the \texttt{CounterToSinkPnP} task, we evaluate on eight new object categories (carrot, ladle, lime, apple, orange, sponge, cucumber, and banana) and new instances of the original object categories with differing size, color, and shape. For the \texttt{CupPnP} task, we replace the red cup with cups of different colors and introduce novel objects. The settings of generalization experiments are detailed in Appendix~\ref{app:generalization}. As shown in Table~\ref{tab:generalization}, the policy trained solely on \real{} achieves a success rate of only 33\% and 10\% on novel objects, whereas the co-trained policy significantly outperforms it with success rates of 50\% and 80\%. The diversity in simulation data contributes to improved generalizability in real-world policy performance. 

\textbf{Co-training with simulation data enhances policy robustness to novel object positions.}
In this experiment, we exclude real demonstrations where the object is placed in the middle of the workspace, retaining only those where the object is positioned along borders or corners of the rectangular sampling region. During evaluation, we place the objects in the center of the sampling region, which are unseen positions in the real demonstrations. The setups are visualized in Appendix~\ref{app:generalization}. In simulation, we still include data with objects distributed uniformly in the rectangular region. As shown in Table~\ref{tab:generalization}, policies co-trained with \DC{} achieve a twice higher success rate compared with the policies trained solely on \real{} for both humanoid and Panda experiments. This result indicates that diverse simulation data substantially improve policy robustness to spatial variations.

\begin{table}[t]
\centering
\begin{tabular}{l|cc|cc}
    \toprule
    \multirow{2}{*}{Data Composition}
      & \multicolumn{2}{c|}{Unseen Objects} 
      & \multicolumn{2}{c}{Unseen Positions} \\
    \cmidrule(lr){2-3}\cmidrule(lr){4-5}
      & Panda & GR-1 & Panda & GR-1 \\
    \midrule
    \texttt{Real}         & 33\% & 10\%   & 11\% & 43\%   \\
    \texttt{Real} + \texttt{DC} & 50\% & 80\%   & 28\% & 100\%   \\
    \bottomrule
\end{tabular}
\caption{\small{
\textbf{Co-training with sim enhances policy generalization across novel objects and novel positions.} We select the \texttt{CounterToSinkPnP} task on Panda and the \texttt{CupPnP} task on the humanoid and evaluate the policies' performance when the object is changed and when the object is placed at unseen positions.
}}
\label{tab:generalization}
\end{table}

\begin{figure}[t]
\center
\includegraphics[width=0.48\textwidth]{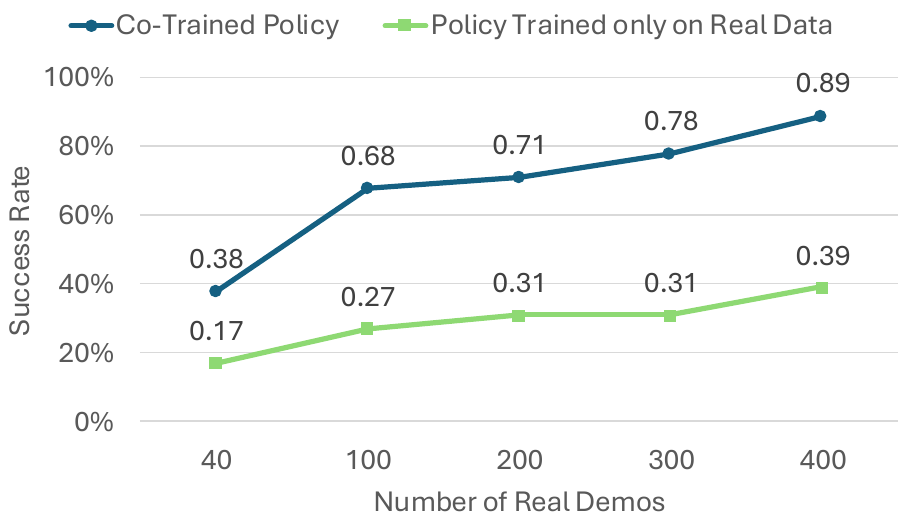}
\caption{\small{
\textbf{Effect of the quantity of real demonstrations.} 
We use a total of 4,000 simulation \texttt{DC} demos and vary the total number of real demos from 40 to 400 on task \texttt{MultiTaskPnP}. The results show that our co-training recipe remains beneficial with larger real datasets.}}
\vspace{-5pt}
\label{fig:scaling_plot}
\end{figure}

\begin{figure}[t]
\center
\includegraphics[width=0.48\textwidth]{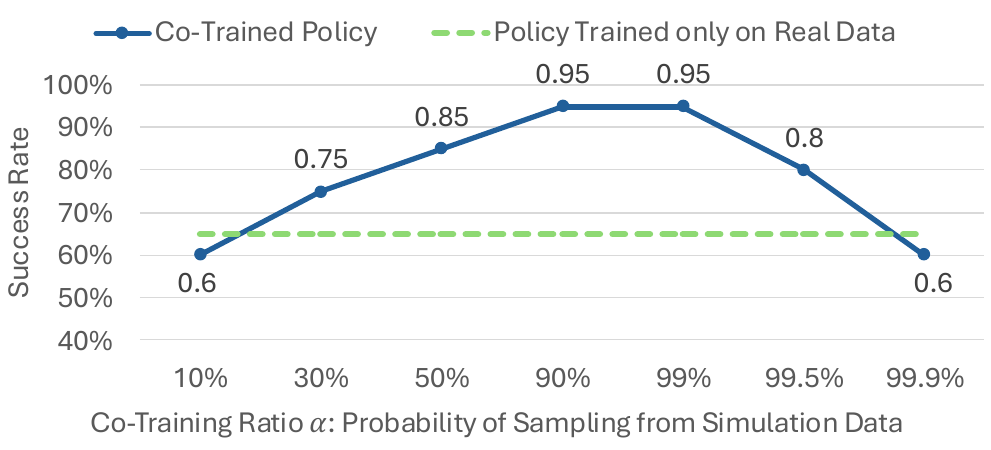}
\caption{\small{
\textbf{Effect of the different co-training ratios.} The co-training ratio, $\alpha$, is the probability of sampling from simulation data in each minibatch. We experiment on the \texttt{CupPnP} task with 20 real demos and 1000 simulation demos from \dc{}. Tuning the co-training ratio is important for the good performance of co-trained policies. 
}}
\label{fig:sim_real_ratio}
\end{figure}

\subsection{Effectiveness of Co-Training in Data-Rich Settings}
\label{subsec:datarich}
Our main results (Section~\ref{subsec:effectiveness}) examine the effectiveness of sim-and-real co-training in single-task settings with different embodiment and scene setups. 
Meanwhile, recent works~\cite{black2024pi_0,kim2024openvla,open_x_embodiment_rt_x_2023} have demonstrated incredible performance by training policies on large-scale, multi-task real-world manipulation datasets. 
This raises the question---can co-training with synthetic data still be beneficial with larger real-robot datasets? 
To evaluate the effectiveness of our co-training approach in a scaled-up setting, we conduct experiments on a new humanoid \texttt{MultiTaskPnP} task. In this task, the robot must pick an object from one container and place it into another, with four different source-target container combinations. 

For each real-world task variation, we construct a corresponding task-aware digital cousin. We train policies using a fixed set of 4,000 \texttt{DC} demonstrations (1,000 per task) while varying the number of real-world demonstrations.
During testing, we evaluate the generalizability of the policy using three unseen objects. The detailed task setup is provided in Appendix~\ref{app:multitaskpnp}.  

As shown in Figure~\ref{fig:scaling_plot}, increasing the number of real-world demonstrations improves the success rate for both real-only and co-trained policies. \textbf{Even with 400 real demonstrations, the co-trained policy consistently outperforms the real-only policy, demonstrating that sim-and-real co-training remains beneficial even in data-rich settings.}

\subsection{Key Elements of Effective Co-Training}
\label{subsec:cotrain_factors}

In the sections above, we demonstrated the effectiveness of sim-and-real co-training with our full-fledged recipe. In this section, we present systematic studies that help identify key elements for successful co-training. These elements include the quantity of real and simulation demonstrations, the co-training ratio, and the importance of camera alignment in digital cousin (\DC{}) environments. Our findings highlight practical strategies for optimizing co-training and improving real-world policy performance.

\textbf{A sufficient number of simulation demonstrations is crucial for effective co-training.}  
First, we analyze the impact of varying the number of simulation demonstrations. We reduce the number of \texttt{DC} demonstrations for the Panda \texttt{CounterToSinkPnP} and GR-1 \texttt{CupPnP} tasks and train policies using \texttt{Real} data combined with the reduced \texttt{DC} data. In the Panda \texttt{CounterToSinkPnP} task, decreasing the number of simulation demonstrations from 10k to 500 causes the success rate to drop from 67\% to 53\%. Similarly, in the GR-1 \texttt{CupPnP} task, reducing the number of simulation demonstrations from 1k to 100 lowers the success rate from 95\% to 75\%. Therefore, having sufficient simulation demonstrations is essential for achieving strong performance in co-trained policies.

\textbf{Tuning the co-training ratio is required for effective co-training.} We investigate the impact of the co-training ratio—the proportion of simulation data used during training—on the \texttt{CupPnP} task (Section~\ref{sec:study-real}). As shown in Figure~\ref{fig:sim_real_ratio}, a 1:1 ratio (50\%) is suboptimal. In our experiments, a co-training ratio of 99\% yielded the best performance. However, pushing the co-training ratio further to values like 99.5\% and 99.9\% resulted in drops in success rate, from 95\% to 60\%. These findings highlight the importance of carefully tuning the co-training ratio to achieve optimal performance.

\begin{figure}[t]
    \centering
    \includegraphics[width=0.48\textwidth]{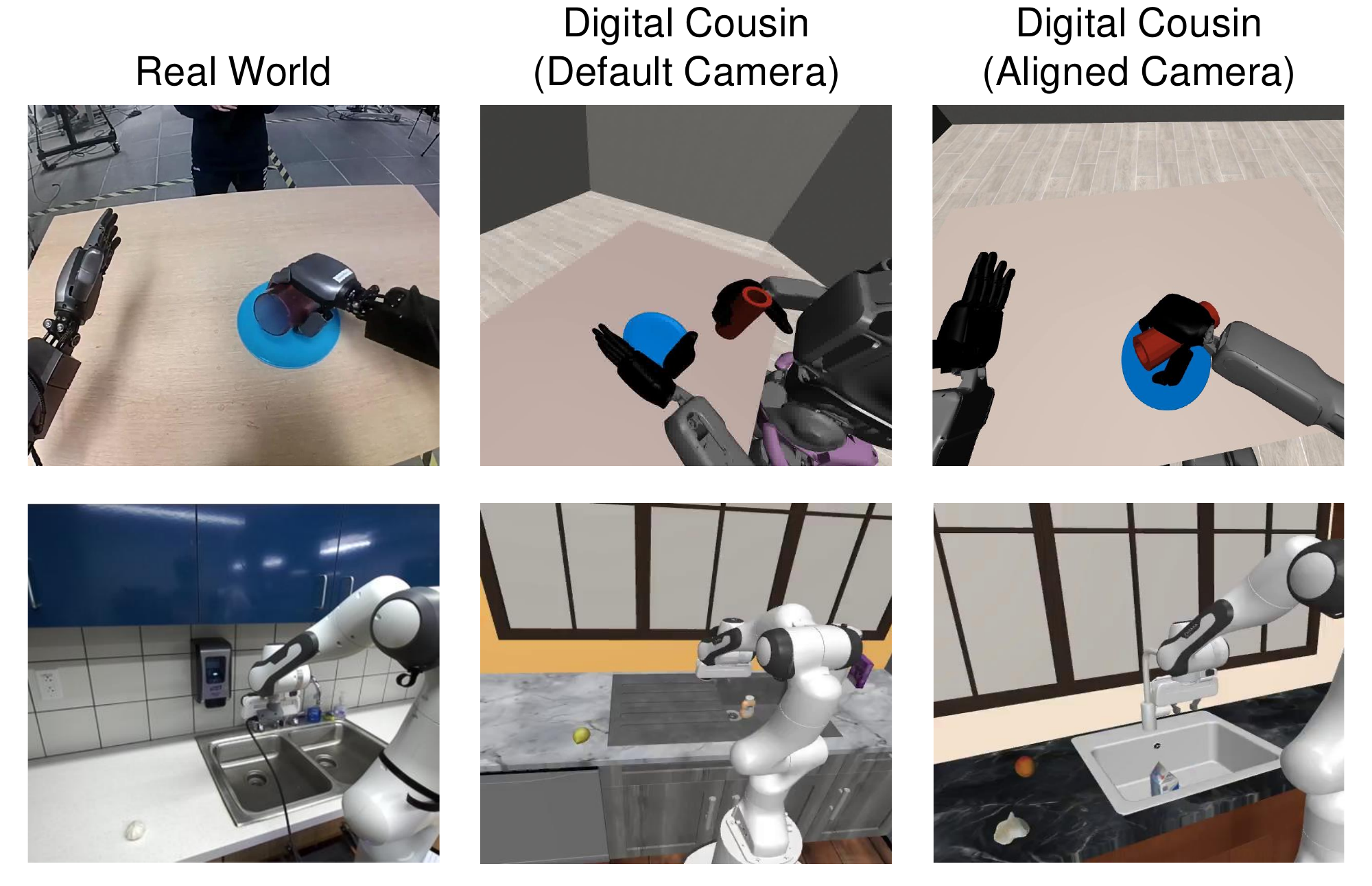}
    \caption{\textbf{Camera alignment visualization.} We visualize the default and aligned camera views of the GR-1 \texttt{CupPnP} task and Panda \texttt{CounterToSinkPnP} task.}
    \vspace{-5pt}
    \label{fig:camera-view}
\end{figure}

\textbf{Camera alignment is critical for successful co-training with task-aware digital cousin data.}  
In \DC{}, we approximate the alignment of the simulation environment's camera with the real-world camera view. To evaluate the importance of this alignment, we render data using the default, unaligned camera view and trained policies on the resulting misaligned simulation data.
The results indicate a significant drop in performance compared to policies co-trained with properly aligned \DC{} data. On the Panda arm \texttt{CounterToSinkPnP} task, the co-training success rate dropped from 67\% to 56\%, while in the GR-1 humanoid \texttt{CupPnP} task, it declined from 95\% to 70\%. We visualize the default and aligned camera views in Figure~\ref{fig:camera-view}. Notably, the aligned camera is not strictly identical to the real-world camera. For example, the camera mounted on the real-world humanoid has a fisheye effect, whereas our aligned simulation camera does not model this distortion. This suggests that while alignment enhances performance, perfect camera alignment is not necessary for effective co-training.

\subsection{A Simple Recipe for Sim-and-Real Co-Training}
\label{subsec:recipe}

Based on our empirical findings, we provide a set of recommendations to help practitioners reap the benefits of co-training with synthetic simulation data.  

\begin{itemize}  
    \item \textbf{Task and scene composition.} The greatest performance gains are observed when co-training with simulation data from task-aware digital cousins, where the task and scene compositions closely mirror those of the real-world setting. Nevertheless, co-training with large multi-task prior simulation data---despite differences in task and scene composition---still provides meaningful benefits.  
    \item \textbf{Object composition and initialization distribution.} Incorporating diverse objects and varying their placements in simulation data helps real-world policies generalize to unseen scenarios.  
    \item \textbf{Alignment between the task-aware digital cousin and the real world.} It is essential that the simulation task shares the same definition and success criteria as its real-world counterpart. Additionally, maintaining similar camera viewpoints between simulation and real-world settings can improve performance, though perfect alignment is not required.  
    \item \textbf{Co-training hyperparameters.} We recommend utilizing a sufficiently large amount of simulation data (ideally, orders of magnitude more than real-world data) and carefully tuning the co-training ratio to optimize performance.  
\end{itemize}

%% file: sections/06_limitations.tex
\section{Limitations}
\label{sec:limitatioins}

Although we conduct systematic studies on several tasks across both a tabletop manipulator and a humanoid robot, most tasks are centered around pick-and-place. Extending our approach to a broader set of manipulation tasks, such as high-precision insertion, and longer-horizon tasks, is left for future work. 
While our co-training recipe consistently improves success rates compared to solely training on real-robot data collection, the policy's performance is still not perfect. Future efforts could look into building on top of this recipe to further boost real-world performance.
Finally, certain real-world tasks---particularly those involving deformable objects and liquids---remain difficult to simulate accurately, inherently limiting the applicability of simulation data. Applying this co-training strategy to such tasks presents a challenge. Future work could explore the use of co-training data produced by video generation models and world models~\cite{nvidia2025cosmos,bruce2024genie,hu2023gaia1} as a way to bridge this gap.

%% file: sections/05_conclusion.tex
\section{Conclusion}
\label{sec:conclusion}

In this work, we systematically investigate how to effectively leverage synthetically generated data from physics simulations to solve real-world, vision-based manipulation tasks. By analyzing key factors that impact the dataset distributions and co-training strategies, we demonstrate that large-scale simulation data can effectively complement real-world data---even in the presence of significant discrepancies---leading to policies that outperform those trained on real-world data alone.

Furthermore, we find that simulation data enhances policy generalization to scenarios not covered in real-world datasets, underscoring its potential for developing more robust and adaptable robotic systems. Our findings highlight the promise of leveraging diverse simulation data to advance generalist robot autonomy.

In addition, we offer a set of practical recommendations for practitioners to harness the benefits of synthetic simulation data without requiring extensive manual effort in constructing or aligning simulation environments. They reinforce the importance of systematically integrating simulation and real-world data.
We hope our insights will inspire future research to further unleash the potential of simulation in building generalizable robot models in the real world.

%% file: sections/appendix.tex
\section{Appendix}

\subsection{Overview}
\label{app:overview}

The Appendix contains the following content:

\begin{itemize}
    \item \textbf{Author Contributions} (Appendix~\ref{app:author}): list of each author's contributions to the paper
    \item \textbf{Real-World Tasks} (Appendix~\ref{app:setup}): details about real-world domains and tasks 
    \item \textbf{Task-Agnostic Simulation Datasets} (Appendix~\ref{app:priordata}): details about task-agnostic simulation environments and datasets 
    \item \textbf{Task-Aware Digital Cousin Datasets} (Appendix~\ref{app:digitalcousins}): details about the tuning process for digital cousin environments and generating large-scale data
    \item \textbf{Data Composition Analysis} (Appendix~\ref{app:datacomposition}): comparing data composition factors across real, digital cousin, and prior datasets
    \item \textbf{Training Details} (Appendix~\ref{app:training}): training algorithm, model architecture, and training protocols 
    \item \textbf{Policy Evaluation} (Appendix~\ref{app:evaluation}): experiment evaluation protocols 
    \item \textbf{Generalization Experiment Details} (Appendix~\ref{app:generalization}): details about the generalization experiments presented in Section~\ref{subsec:generalization}
    \item \textbf{MultiTaskPnP Task Setup} (Appendix~\ref{app:multitaskpnp}): details about the multi-task experiment setup presented in Section~\ref{subsec:datarich}
    \item \textbf{Improving Visual Realism with Vid2Vid} (Appendix~\ref{app:vid2vid}): additional experiments on bridging the visual gap between simulation and real-world data with Vid2Vid techniques
    \item \textbf{Training CloseDoor with More Demos} (Appendix~\ref{app:closedoorextrademos}): additional experiments to improve performance on real-world CloseDoor task
    \item \textbf{FAQ} (Appendix~\ref{app:faq}): additional topics
\end{itemize}

\subsection{Author Contributions}
\label{app:author}
\begin{itemize}
    \item \noindent \textbf{Algorithm design:} Soroush Nasiriany, Zhenyu Jiang, Lawrence Yunliang Chen, Abhiram Maddukuri, Ajay Mandlekar, Yuke Zhu
    \item \noindent \textbf{Experiments (Panda Kitchen):} Abhiram Maddukuri, Soroush Nasiriany
    \item \noindent \textbf{Experiments (Humanoid Tabletop):} Zhenyu Jiang, Lawrence Yunliang Chen, Yuqi Xie, Zu Wang
    \item \noindent \textbf{Infrastructure support:} Yu Fang, Wenqi Huang, Nikita Chernyadev, Zhenjia Xu, Soroush Nasiriany
    \item \noindent \textbf{Digital release:} Lawrence Yunliang Chen, Zhenyu Jiang
    \item \noindent \textbf{Technical advice:} Ken Goldberg, Scott Reed
    \item \noindent \textbf{Project leads:} Yuke Zhu, Linxi Fan, Ajay Mandlekar
\end{itemize}

\subsection{Real-World Tasks}
\label{app:setup}
\noindent \textbf{Panda Kitchen.} For our Panda Kitchen tasks, we use the DROID \cite{khazatsky2024droid} setup and make some modifications. We use the controller from Deoxys~\cite{zhu2022viola}, which supports OSC-based end effector control~\cite{khatib1995osc}, and opt to keep the original parallel-jaw Panda gripper instead of the Robotiq gripper. We use two side-view third-person cameras and an eye-in-hand camera.

\begin{figure}[t]
    \centering
    \includegraphics[width=\linewidth]{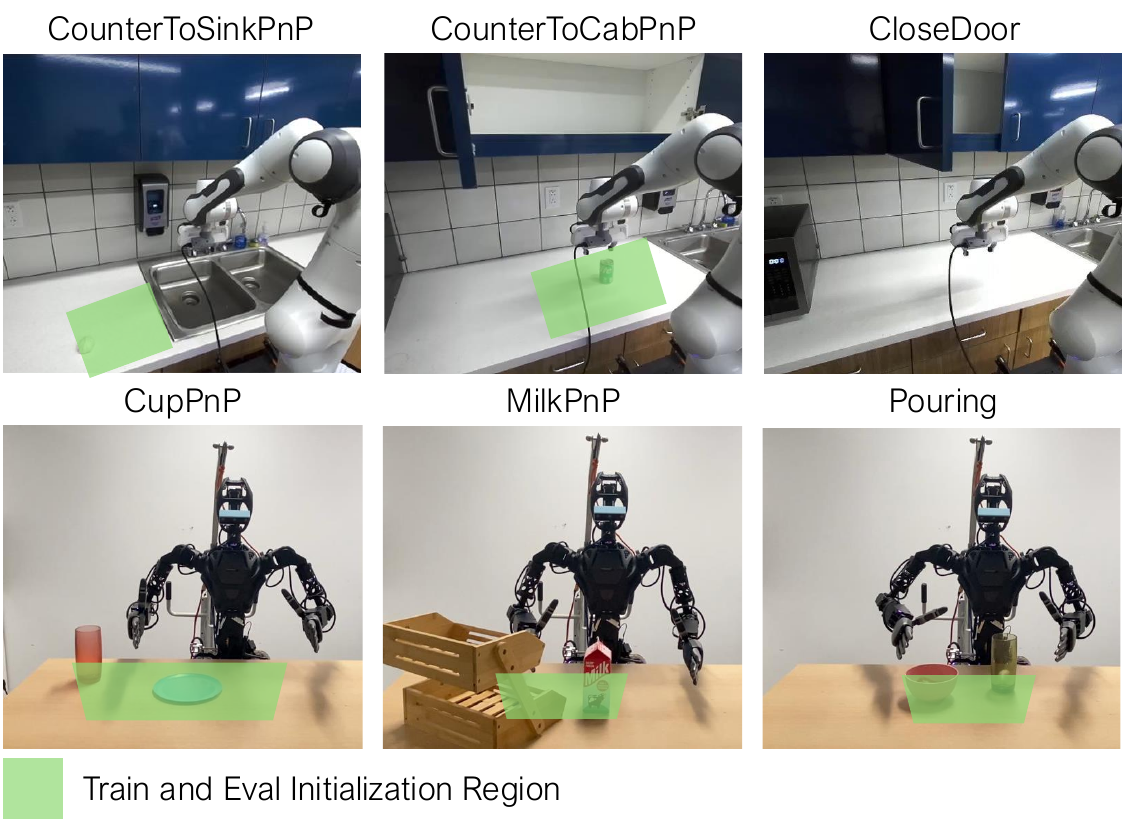}
    \caption{\textbf{Visualization of start states for our experiments.} We visualize the starting states of the Panda and GR-1 for our experiments, including initialization regions for our pick-and-place tasks.}
    \label{fig:init-vis}
    \vspace{-10pt}
\end{figure}

For each task, we collect data via teleoperation using a SpaceMouse. The robot starts from a fixed initial position and joint configuration, with camera poses remaining constant across each demo. For the \texttt{CounterToSinkPnP} task, we uniformly sample object placements from the rectangular counter region to the left of the sink. For the \texttt{CounterToCabPnP} task, we uniformly sample object placements from the rectangular counter region below the cabinet. Finally, in the \texttt{CloseDoor} task, we uniformly sample the initial angle of the open door from the interval [$85^\circ$, $115^\circ$]. Refer to Figure ~\ref{fig:init-vis} for a visualization of the start states and sampling regions for each task.\\
\\
\noindent \textbf{Humanoid Tabletop.} We use a Fourier GR-1 robot with the default 6-DoF dexterous hands. The joints of the robot's lower body and waist are locked, while the two arms and hands are activated. During data collection, a human operator wears a pair of MANUS gloves with a VIVE tracker positioned on the back of each hand to capture the finger pose and wrist pose, respectively. We implement an inverse kinematic (IK) controller based on the mink framework~\cite{Zakka_Mink_Python_inverse_2024} to control the robot during teleoperation, where the human input consists of wrist pose and finger joint commands. During policy learning and deployment, we use the joint positions computed by the IK solver as the action space. Additionally, an OAK-D camera mounted on the robot's head provides egocentric visual input for the system. Only one of the stereo RGB images (and not the depth) is used for the policy, and no additional third-person view is used.

For each of the three single tasks---\texttt{CupPnP}, \texttt{MilkPnP}, and \texttt{Pouring}, we collect 20 demonstrations in the real world. The initial pose of the robot varies around a standard reset pose according to the human teleoperator's pose. Only one set of objects is used for each task, but the object positions are randomized uniformly in the region as visualized in Figure~\ref{fig:init-vis}.

For the multi-task setting (\texttt{Cuttingboard2Basket}, \texttt{Cuttingboard2Pan}, \texttt{Mat2Basket}, and \texttt{Plate2Bowl}), we additionally select multiple object instances during data collection. See Appendix~\ref {app:multitaskpnp} for details.

\subsection{Task-Agnostic Simulation Datasets}
\label{app:priordata}
\noindent\textbf{Panda Kitchen.} We use a subset of the 72k trajectories from the RoboCasa dataset \cite{robocasa2024}. Specifically, we exclude the \texttt{OpenDrawer}, \texttt{CloseDrawer}, \texttt{TurnOnStove}, and \texttt{TurnOffStove} tasks from the original dataset due to lack of visibility introduced by camera pose alignment. This results in 60k trajectories over the following 20 tasks. We cite the task descriptions from RoboCasa here:
\begin{enumerate}
    \item \texttt{PickPlaceCounterToCabinet}: Pick an object from the counter and place it inside the cabinet.

    \item \texttt{PickPlaceCabinetToCounter}: Pick an object from the cabinet and place it on the counter.

    \item \texttt{PickPlaceCounterToSink}: Pick an object from the counter and place it in the sink.

    \item \texttt{PickPlaceSinkToCounter}: Pick an object from the sink and place it on the counter area next to the sink.

    \item  \texttt{PickPlaceCounterToMicrowave}: Pick an object from the counter and place it inside the microwave.

    \item \texttt{PickPlaceMicrowaveToCounter}: Pick an object from inside the microwave and place it on the counter. 
    
    \item \texttt{PickPlaceCounterToStove}: Pick an object from the counter and place it in a pan or pot on the stove
    
    \item \texttt{PickPlaceStoveToCounter}: Pick an object from the stove (via a pot or pan) and place it on the counter.
    
    \item \texttt{OpenSingleDoor}: Open a microwave door or a cabinet with a single door.
    
    \item \texttt{OpenDoubleDoor}: Open a cabinet with two opposite-facing doors.

    \item \texttt{CloseSingleDoor}: Close a microwave door or a cabinet with a single door.
    
    \item \texttt{CloseDoubleDoor}: Close a cabinet with two opposite-facing doors.
      
    \item \texttt{TurnOnSinkFacuet}: Turn on the sink faucet to begin the flow of water.
    
    \item \texttt{TurnOffSinkFaucet}: Turn off the sink faucet to stop the flow of water.
    
    \item \texttt{TurnSinkSpout}: Turn the sink spout.
    
    \item \texttt{CoffeePressButton}: Press the button on the coffee machine to pour coffee into the mug.

    \item \texttt{TurnOnMicrowave}: Turn on the microwave by pressing the start button.

    \item \texttt{TurnOffMicrowave}: Turn off the microwave by pressing the stop button.

    \item \texttt{CoffeeSetupMug}: Pick the mug from the counter and insert it into the coffee machine mug holder area.

    \item \texttt{CoffeeServeMug}: Remove the mug from the coffee machine mug holder and place it on the counter.
\end{enumerate}
\noindent\textbf{Humanoid Tabletop.} The tabletop humanoid datasets comprise 10 tasks in the RoboCasa simulation framework:

\begin{enumerate}
    \item \texttt{CounterToPlate}: Pick up an object from the tabletop and place it on the nearby plate.
    \item \texttt{PnPAppleToPlate}: Pick up the apple from the tabletop and place it on the nearby plate.
    \item \texttt{PnPCanToBowl}: Pick up the can from the tabletop and place it in the nearby bowl.
    \item \texttt{PnPMugToPlate}: Pick up the mug from the tabletop and place it on the nearby plate. 
    \item \texttt{PnPFruitPlacement}: Pick up the fruit from the tabletop and place it on the nearby plate.
    \item \texttt{PnPKettleToPlate}: Pick up the kettle from the tabletop and place it on the nearby plate.
    \item \texttt{PnPMilkPlateToPlate}: Pick up the milk carton located on a plate and transfer it to another plate.
    \item \texttt{PnPMilkToBasket}: Pick up the milk carton from the tabletop and place it in the nearby basket.
    \item \texttt{PnPPlateToPlate}: Pick the object located on a plate and transfer it to another plate.
    \item \texttt{PnPVegetableBowlToPlate}: Pick the vegetable located inside a bowl and transfer it to a plate.
\end{enumerate}

\subsection{Task-Aware Digital Cousin Datasets}
\label{app:digitalcousins}

\noindent\textbf{Panda Kitchen.}
We design a \textit{digital cousin} for each of the three tasks in the Panda Kitchen. Among the set of data composition factors outlined in Section~\ref{subsec:cotrain_factors}, we focus on approximately aligning the following factors: Task composition, Object composition, and Initialization distribution. See Tables \ref{tab:panda-kitchen-datasets} and \ref{tab:countertosinkpnp-datasets} for a summary of the alignments.
\begin{itemize}
    \item \textbf{Task composition:} We align our digital cousin to contain the same task semantics and motions as in the real world.  This involves using identical language instructions to those in the real task and approximately matching the initial joint configurations and positions to ensure similar trajectory distributions.

    \item \textbf{Object composition:} We design the digital cousin for our \texttt{Pick-and-Place} tasks to predominantly include object categories present in the real-world setting. Specifically, in the \texttt{CounterToSinkPnP DC}, 9/10 object categories overlap with those used in real-world tasks. Similarly, in the \texttt{CounterToCabPnP DC}, 8/10 object categories are shared with the real-world setting.

    \item \textbf{Initialization distribution:} We approximately align starting joint states, initial positions, and object initialization regions for our digital cousins. Specifically, we update the default starting position of the Panda to be farther away from the counterspace and increase the starting height to better match the starting position in the real world. We also reduce the area of the default sampling region to better match the sampling region in the real-world setting. 
    
\end{itemize}

For our \texttt{CounterToSinkPnP DC}, \texttt{CounterToCabPnP DC}, and \texttt{CloseDoor DC}, we generate 10,000 synthetic trajectories each from an initial source of 100 human demonstrations using MimicGen \cite{mandlekar2023mimicgen}. We further diversify these demonstrations by rendering them using combinations of the 100 floor, 100 wall, 100 cabinet, and 100 counter AI-generated textures provided by RoboCasa \cite{robocasa2024}.
\\ \\
\noindent\textbf{Humanoid Tabletop.}
Similarly, we design one digital cousin (DC) for each of the three tasks in the Humanoid Tabletop domain. We focus on the following factors when designing the DC: Task composition, Object composition, and Initialization distribution.
\begin{itemize}
    \item \textbf{Task composition}: we set up the scene and the success check to make the simulation task semantically similar to the real-world task. During the simulation source demo collection, we try to mimic the motion pattern of the real-world demos. We add a small random offset (uniform distribution between $[-0.2, 0.2]$) to the initial joint distribution as real demos have different initial joints.
    \item \textbf{Object composition}: since we have a fixed set of objects in the real-world, we use a smaller set of objects in DC for the GR-1 humanoid, as noted in Table~\ref{tab:humanoid-tabletop-datasets}.
    \item \textbf{Initialization distribution}: we align object initialization regions for our DC with the real-world counterparts.
\end{itemize}

For our \texttt{CupPnP}, \texttt{MilkPnP}, and \texttt{Pouring} digital cousins, we generate 1000 synthetic demonstrations each from an initial source of 10 human demonstrations using DexMimicGen \cite{jiang2024dexmimicen}. We instantiate the \texttt{MilkPnP} and \texttt{Pouring} digital cousins in RoboCasa, and we render them with 10 table textures.

\subsection{Data Composition Analysis}
\label{app:datacomposition}

We provide a detailed comparison of the real-world, task-aware digital cousin and task-agnostic prior datasets in Table~\ref{tab:panda-kitchen-datasets} for the Panda Kitchen domain and Table~\ref{tab:humanoid-tabletop-datasets} for the Humanoid Tabletop domain.

We also add a detailed comparison of additional factors such as camera positions and initialization parameters for the \texttt{CounterToSinkPnP} task in Table~\ref{tab:countertosinkpnp-datasets}.

\begin{table}[]
    \centering
    \resizebox{\linewidth}{!}{
    \begin{tabular}{l|cccccc}
    \toprule
    & Dataset & Obj. Cat. & Obj. Ins. & Scenes & Demos \\ \midrule
    \multirow{2}{*}{\texttt{Real}} & CounterToSinkPnP & 9 & 13 & 1 & 50 \\
                                   & CounterToCabPnP & 8 & 12 & 1 & 50 \\
                                   & CloseDoor & N.A. & N.A. & 1 & 50 \\ \midrule
    \multirow{2}{*}{\texttt{DC}}   & CounterToSinkPnP & 10 & 67 & 100 & 10k \\
                                   & CounterToCabPnP & 10 & 67 & 100 & 10k \\
                                   & CloseDoor & N.A. & N.A. & 100 & 50 \\ \midrule
    \texttt{Prior}                 & RoboCasa (20 tasks) & 70 & 381 & 100 & 60k \\ \bottomrule
    \end{tabular}}
    \caption{\small{\textbf{Aggregate dataset statistics of the Panda Kitchen domain.} We outline the number of object categories, object instances, scenes, and number of demonstrations for the real, digital cousin (DC), and task-agnostic prior (Prior) datasets.}}
    \label{tab:panda-kitchen-datasets}
    \vspace{-12pt}
\end{table}

\begin{table}[]
    \centering
    \resizebox{0.90\linewidth}{!}{
    \begin{tabular}{l|cccccc}
    \toprule
                                   & Dataset & %
                                   Obj. Cat. & Table Tex. & Demos \\ \midrule
    \multirow{3}{*}{\texttt{Real}} & CupPnP & 1 & 1 & 20 \\
                                   & MilkPnP & 1 & 1 & 20 \\
                                   & Pouring & 1 & 1 & 20 \\ \midrule
    \multirow{2}{*}{\texttt{DC}}   & CupPnP & 1 & 1 & 1k \\
                                   & MilkPnP & 3 & 10 & 1k \\
                                   & Pouring & 2 & 10 & 1k \\ \midrule
    \texttt{Prior}                 & RoboCasa (10 tasks) & 66 & 10 & 10k \\ \bottomrule
    \end{tabular}}
    \caption{\small{\textbf{Aggregate dataset statistics of the Humanoid Tabletop domain.} We outline the number of object categories, table textures, and number of demonstrations for the real, digital cousin (DC), and task-agnostic prior (Prior) datasets. Note that for \texttt{CupPnP}, we randomize the color of the object, although we are using a single object category.}}
    \label{tab:humanoid-tabletop-datasets}
    \vspace{-12pt}
\end{table}

\begin{table*}[]
    \centering
    \resizebox{0.95\linewidth}{!}{
    \begin{tabular}{l|cccccc}
    \toprule
    & Left Cam Pos. & Right Cam Pos. & Wrist Cam Pos. & Init. Robot Joints & Init. Robot Base Pos.  & Obj. Init. Reg. \\ \midrule
    \texttt{Real} & [-0.35, 0.49, 0.70] & [-0.35, -0.42, 0.72] & [-0.029, 0, 0.05] & (0.09, -0.20, -0.02, -2.47, -0.01, 2.30, 0.85) & [-18, -26, 0] & 27x23\\
    \texttt{DC} & [-0.35, 0.49, 0.70] & [-0.35, -0.42, 0.72] & [-0.029, 0, 0.05] & (0.09, -0.20, -0.02, -2.47, -0.01, 2.30, 0.85) & [0, -32, -4] & 27x27 \\
    \texttt{Prior} & [-0.35, 0.49, 0.70] & [-0.35, -0.42, 0.72] & [-0.029, 0, 0.05] & (-0.02,
                    -1.03,
                    -0.02,
                    -2.28,
                    0.04,
                    1.52,
                    0.70) & [0, -20, -22] & 30x40 \\ 
    \bottomrule
    \end{tabular}}
    \caption{\small{\textbf{Additional dataset statistics for \texttt{CounterToSinkPnP} task.} All units except for initial robot joints are in centimeters. Camera positions are (x,y,z) coordinates relative to the robot base. Initial robot base positions are (x, y, z) coordinates relative to the middle edge of the sink. Object initialization region is (depth x width)}}
    \label{tab:countertosinkpnp-datasets}
    \vspace{-10pt}
\end{table*}

\subsection{Training Details}
\label{app:training}
\noindent\textbf{Panda Kitchen.} We adopt the open-source Diffusion Policy implementation from Chi et al.~\cite{chi2023diffusion}. We use the transformer-based variant with ResNet visual encoders. The policy takes three $128 \times 170$ image views and robot proprioception information (end effector position and rotation, gripper joint values), and outputs a 7-DoF action for delta end-effector control and gripper action. We modify the hyperparameters to use a larger transformer network and a larger batch size of 256. We also add language conditioning to facilitate training on diverse multi-task data; we encode language using the CLIP sentence encoder, and we add FiLM conditioning layers~\cite{perez2017film} to the vision encoder. We set a default co-training ratio of $0.10$ for real-world data and $0.90$ for simulation data. We do this as we have a significantly higher quantity of simulation demonstrations than real-world demonstrations.

As mentioned in Section~\ref{subsec:cotraining_ratio_prelim}, with a training batch of size $B$, on average, $\alpha \cdot B$ samples are drawn from the simulation dataset $\mathcal{D}_{\text{sim}}$, and $(1-\alpha) \cdot B$ samples are drawn from the real-world dataset $\mathcal{D}_{\text{real}}$. To enforce this, we reweight each data sample such that $P[(o_i, a_i) \in \mathcal{D}_{\text{sim}}] = \alpha$ and $P[(o_i, a_i) \in \mathcal{D}_{\text{real}}] = 1-\alpha$, where $(o_i, a_i)$ denotes an observation-action pair within the batch. We achieve this by first normalizing the weight of each sample according to the size of its dataset, and then multiplying the normalized weight by $\alpha$ if the sample belongs to $\mathcal{D}_{\text{sim}}$ and by $1-\alpha$ if it belongs to $\mathcal{D}_{\text{real}}$.

\noindent\textbf{Humanoid Tabletop.} Since there is no ambiguity in terms of objects to manipulate, we use Diffusion Policy (DP) without language conditioning and train one policy for each task. We use the DP implementation from UMI~\cite{chi2024universal} with Vision Transformers~\cite{dosovitskiy2020image} as vision encoders and UNet~\cite{ronneberger2015u} as the diffusion backbone. The input observation contains an RGB image from a first-person-view camera and joint position observations. The output action is the target joint positions of the arms and the dexterous hands. Figure~\ref{fig:sim_real_ratio} shows that co-training ratios $\alpha$ of 0.9 and 0.99 are optimal, and we use 0.99 for all other experiments.

\subsection{Policy Evaluation}
\label{app:evaluation}
We use similar evaluation protocols for the Franka Panda arm and the GR-1 humanoid. We evaluate the model at three checkpoints during training, spaced at equal intervals. At each checkpoint, we assess the model’s success rate and use the highest success rate among these evaluations as the final result.

\noindent\textbf{Panda Kitchen.} For both the \texttt{CounterToSink} and \texttt{CounterToCab} tasks, we perform four random placements on the counter per object and record the overall average success rate for all object categories. For the \texttt{CloseDoor} task we sample 10 random joint angles between [$85^\circ$, $115^\circ$] and average the success rate across all sampled joint values; this results in 36, 32, and 10 total rollouts per checkpoint for the \texttt{CounterToSink}, \texttt{CounterToCab}, and \texttt{CloseDoor} respectively. For the \texttt{CounterToSink}, we record a success if the policy picks up the object and places it in the left or right basin of the sink, and record a failure otherwise. For the \texttt{CounterToCab}, we record a success if the policy picks up the object and securely places it in the cabinet, and record a failure otherwise. Finally, for the \texttt{CloseDoor} task, we record a success if the door's joint angle is less than $5^\circ$ and record a failure otherwise.

The initialization regions and starting conditions (robot joint states, robot positions, object sampling regions) at test time are the same as during training. See Figure~\ref{fig:init-vis} for a visualization of the starting state and initialization regions for the tasks. For \texttt{CounterToCabPnP} and \texttt{CounterToSinkPnP}, we evaluate on seen objects, and for \texttt{CloseDoor}, we evaluate on the same door and use the same range of joint angles, [$85^\circ$, $115^\circ$], from the training data.

\noindent\textbf{Humanoid Tabletop.} For each task, we evaluate the policy performance using the same objects and position distributions. For each checkpoint, we evaluate 20 different initial positions for the \texttt{CupPnP} task and 10 for the \texttt{MilkPnP} and \texttt{Pouring} tasks. We report the highest success rates among the three checkpoints for each training run. We consider partial successes, where a successful pick is counted as 0.5.

\begin{figure}[t]
    \centering
    \includegraphics[width=\linewidth]{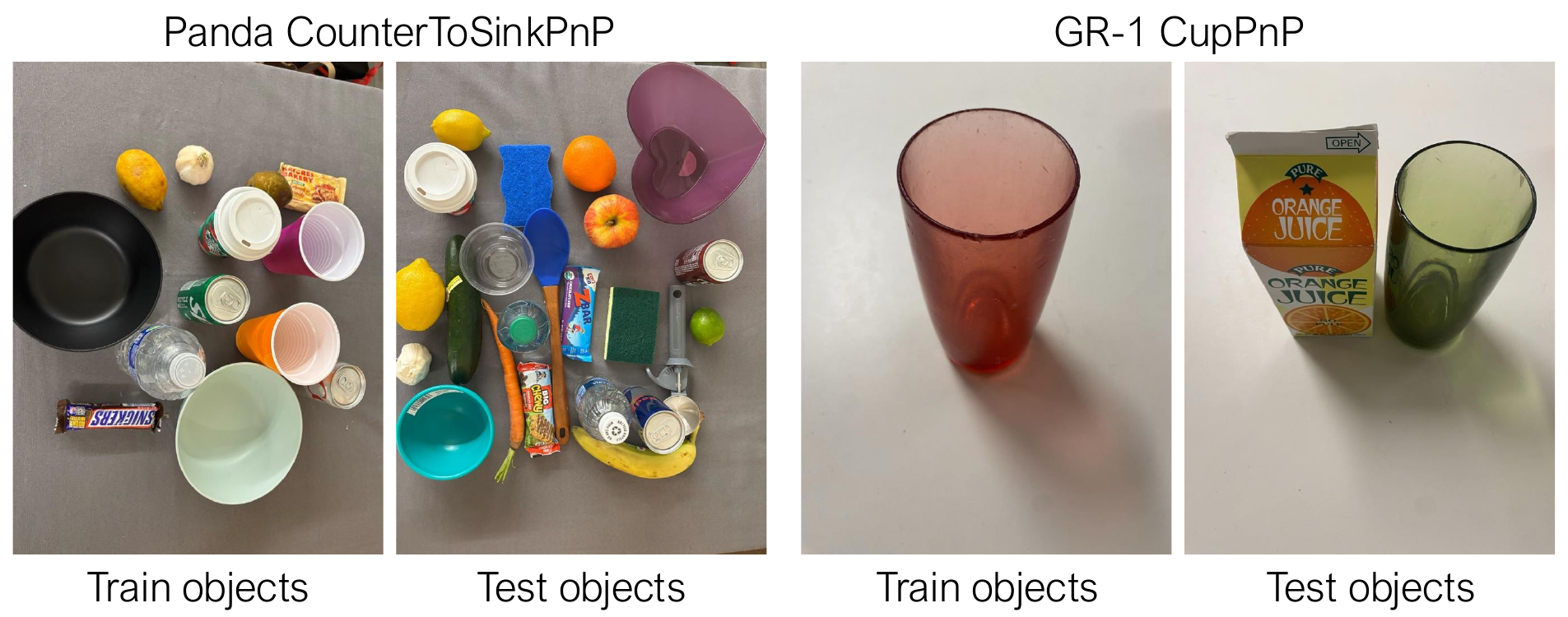}
    \caption{\textbf{Visualization of novel object experiment settings.} We show the picture of train objects and test objects of the generalization experiment conducted on Panda \texttt{CounterToSinkPnP} and GR-1 \texttt{CupPnP} tasks.}
    \label{fig:novel-obj-vis}
    \vspace{-12pt}
\end{figure}

\begin{figure}[t]
    \centering
    \includegraphics[width=\linewidth]{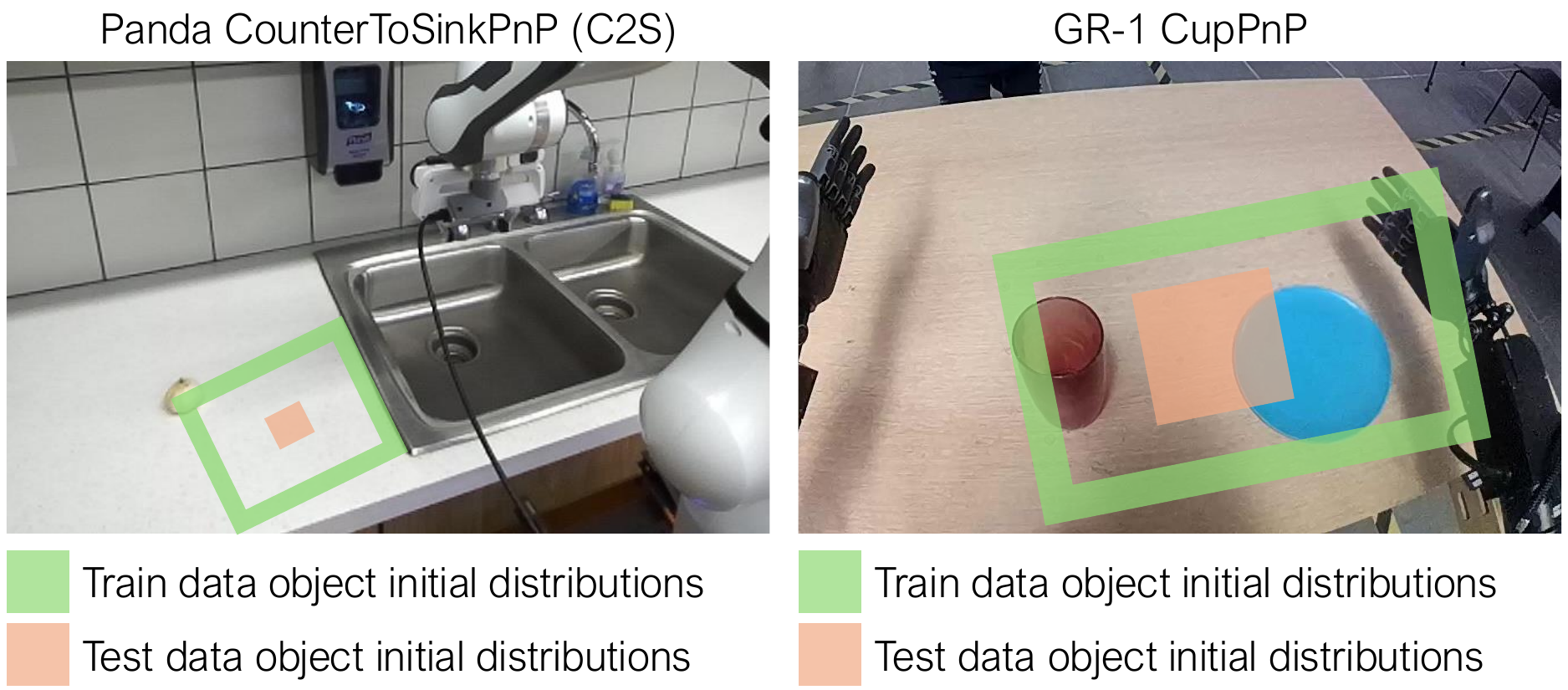}
    \caption{\textbf{Visualization of novel position experiment settings.} We visualize the train and test object initialization range of the generalization experiment conducted on Panda \texttt{CounterToSinkPnP} and GR-1 \texttt{CupPnP} tasks.}
    \label{fig:novel-pos-vis}
    \vspace{-12pt}
\end{figure}

\subsection{Generalization Experiment Details}  
\label{app:generalization}

This section visualizes the settings of the generalization experiments described in Section~\ref{subsec:generalization}. Figure~\ref{fig:novel-obj-vis} illustrates the training and testing objects used in the generalization experiments for the Panda \texttt{CounterToSinkPnP} and GR-1 \texttt{CupPnP} tasks. The default setup for the Panda \texttt{CounterToSinkPnP} task involves manipulating a diverse set of objects, whereas the GR-1 \texttt{CupPnP} task primarily uses the same red cup. Consequently, we observe a much greater performance improvement in the \texttt{CupPnP} task when employing sim-and-real co-training, as in Table~\ref{tab:generalization}.  

Figure~\ref{fig:novel-pos-vis} visualizes the training and testing object initialization ranges for the generalization experiments conducted on the Panda \texttt{CounterToSinkPnP} and GR-1 \texttt{CupPnP} tasks. In the training data, objects are always initialized at the borders of the workspace, while testing is performed with objects placed at the center of the workspace.

\subsection{MultiTaskPnP Task Setup}
\label{app:multitaskpnp}
In the multi-task experiment on the humanoid, we consider the following four tasks:
\begin{enumerate}
    \item \texttt{Cuttingboard2Basket}: Pick up an object from the cutting board and place it into the nearby basket.
    \item \texttt{Cuttingboard2Pan}: Pick up an object from the cutting board and place it into the nearby pan.
    \item \texttt{Mat2Basket}: Pick up an object from the mat and place it into the nearby basket.
    \item \texttt{Plate2Bowl}: Pick up an object from the plate and place it into the nearby bowl. 
\end{enumerate}

\begin{figure}[t]
    \centering
    \includegraphics[width=\linewidth]{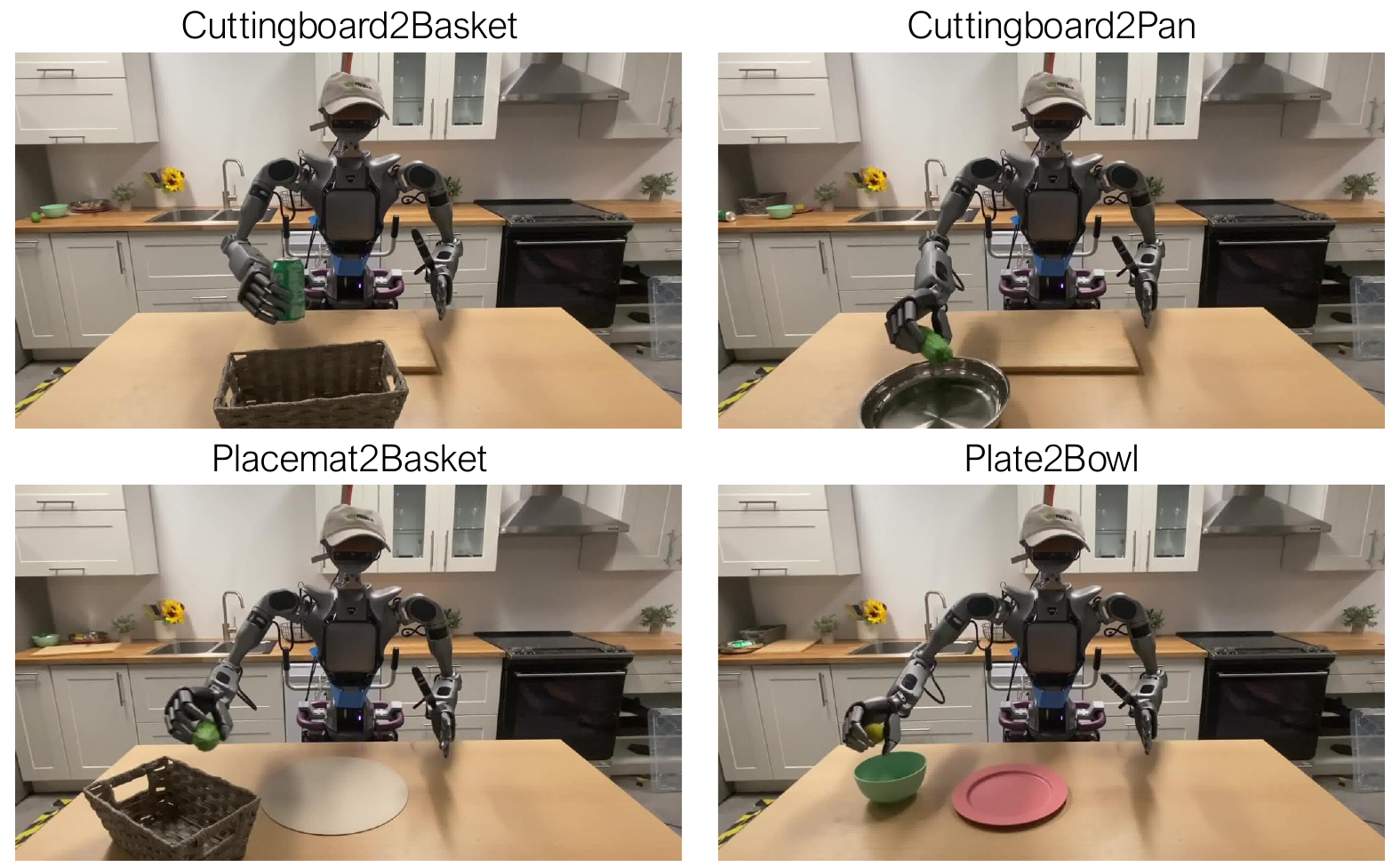}
    \caption{\textbf{\texttt{MultiTaskPnP} visualization.} We show the real-world scene setup of the four tasks in \texttt{MultiTaskPnP}.}
    \label{fig:4DC}
    \vspace{-10pt}
\end{figure}

\begin{figure}[t]
    \centering
    \includegraphics[width=\linewidth]{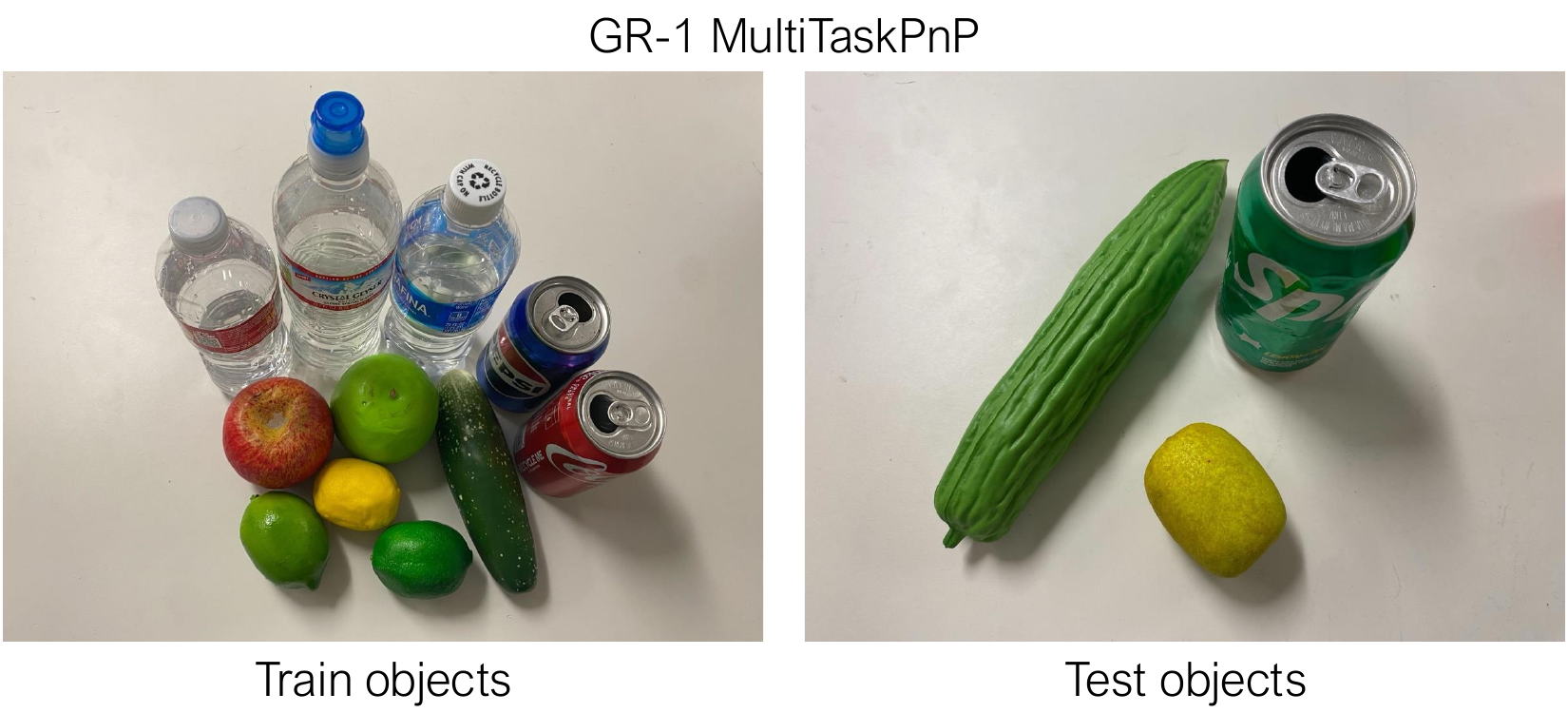}
    \caption{\textbf{\texttt{MultiTaskPnP} train and test objects.} We show the real-world train and test objects we used for GR-1 \texttt{MultiTaskPnP}.}
    \label{fig:4DC-obj}
    \vspace{-10pt}
\end{figure}

The tasks are visualized in Figure~\ref{fig:4DC}. The train and test objects we used are visualized in Figure~\ref{fig:4DC-obj}. We collect 100 demonstrations for each task in the real world, where we randomly select an object and an initial location for each trajectory. At test time, we choose unseen instances of the same object category as the test objects, but the positions are within the demonstration distribution.

We create a digital cousin for each of the four tasks, using target objects and containers of the same category in simulation. Additionally, we align the initial distribution of object positions to approximate the distribution observed in real-world demonstrations. Once the simulation environments are set up, we collect 10 human demonstrations and generate 1000 simulation demonstrations per task using DexMimicGen~\cite{jiang2024dexmimicen}, resulting in a total of 4000 \texttt{DC} demonstrations.  

We train a single policy across all four tasks, where the policy implicitly conditions on the image observation to determine the current task. For evaluation, we select three unseen objects---can, lemon, and cucumber (see Figure~\ref{fig:4DC-obj} right)---and assess performance across three different initial positions for each object in each task, yielding a total of 36 evaluations across the four tasks. We report the average performance across all tasks, considering partial successes, where a successful pick is counted as 0.5.

In Figure~\ref{fig:scaling_plot}, we compare policies trained on \texttt{Real + DC}, where we use 4,000 total \texttt{DC} demonstrations and vary the number of real-world demonstrations, with 10, 25, 50, 75, and 100 real demonstrations per task. Results show that the co-trained policies consistently outperform the real-only policies.

We also compare co-training with different simulation data in this multi-task setting, similar to the single-task setting we show in Section~\ref{subsec:effectiveness}. We use 200 real demonstrations (50 per task), 4000 \texttt{DC} demos (1,000 per task), and 10,000 \texttt{Prior} demos (1,000 per task). Similar to Table~\ref{tab:main_result}, we see that the policy trained on the combination of \texttt{Real}, \texttt{DC}, and \texttt{Prior} data performs the best, getting a success rate of 75.7\%, followed by \texttt{Real + DC} (70.8\%) and \texttt{Real + Prior} (68.8), both of which significantly outperform the \texttt{Real} policy (30.6\%).

\begin{figure*}
    \centering
    \includegraphics[width=0.9\linewidth]{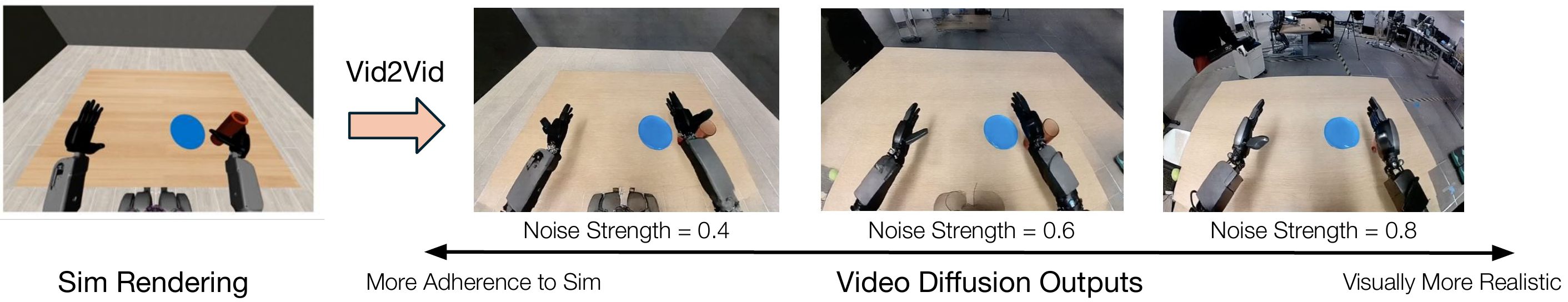}
    \caption{\small{\textbf{Examples of the Video2Video model outputs with different noise strength.} \textbf{Left:} An example video frame from the simulation data. \textbf{Right:} The corresponding frames generated by the trained video diffusion model by initializing the noises with the simulation video with different noise strengths. By setting different values for the noise strength parameter, we can control how much noise is added and from which timestep the model starts diffusion.}}
    \label{fig:vid2vid_examples_noise_strength}
    \vspace{-5pt}
\end{figure*}

\subsection{Improving Visual Realism with Vid2Vid}\label{app:vid2vid}

Given our observation that incorporating task-aware digital cousins, such as camera viewpoint alignment, enhances performance, we pose a follow-up question: \textit{Would improving visual realism by making simulation rendering closer to reality enhance co-training?}

Several studies~\cite{bousmalis2018using, rao2020rl, ho2021retinagan, scheikl2022sim, liu2023digital} have suggested that training generative models (such as CycleGAN) on real and simulated images, and applying them to simulation images to make them visually more realistic, improves sim-to-real policy transfer. Inspired by this, we conduct experiments to assess the extent to which improving the visual realism of simulation data enhances policy performance.

Specifically, we fine-tune CogVideo-X~\cite{yang2024cogvideox}, a state-of-the-art video diffusion model, on \texttt{CupPnP} real demonstration videos in the Text2Video modality. This enables the model to generate realistic-looking videos of a robot performing the task from pure Gaussian noise. 

To adapt the model for style transfer on simulation videos—preserving object positions while enhancing visual realism—we adopt a simple strategy: instead of generating videos from pure noise, we introduce noise into reference simulation videos and use them as initialization for video diffusion. By adjusting the noise strength parameter, we control the extent of noise added and the diffusion starting timestep (see Figure~\ref{fig:vid2vid_examples_noise_strength} for examples). Lower noise levels retain more of the original simulation textures, producing outputs closer to the inputs, while higher noise levels result in more realistic appearances but may distort object poses. 

We set the noise strength to 0.6, as lower values yield outputs too similar to simulation textures, whereas higher values cause excessive deviations from object positions, rendering the original action labels from simulation data inapplicable.

We then conduct co-training experiments in the \texttt{Real} + \texttt{DC} setup on the \texttt{CupPnP} task, comparing policy performance with and without Vid2Vid augmentations on DC data under different numbers of sim and real demos similar to Figure~\ref{fig:scaling_plot}.

\textbf{We find that improved visual realism is particularly beneficial in low-data regimes and provides a modest improvement in overall policy performance.} Experiment results, reported in Table~\ref{tab:vid2vid}%
, show that Vid2Vid-enhanced visual realism leads to a 5–10\% average improvement in policy performance, with the most significant benefits occurring when the number of simulation or real-world trajectories is low. When sufficient real-world demonstrations are available, simulation data plays a minor role. Conversely, when a large and diverse set of simulation data is available, the importance of visual realism diminishes.

These findings highlight the potential of the role of generative models as part of the synthetic data. While our approach utilizes a renderer to provide reference videos for the video diffusion model, future work could explore synthetic generation without reliance on the graphics renderer.

\begin{table}[]
    \centering
    \resizebox{\linewidth}{!}{
    \begin{tabular}{l|ccccc}
    \toprule
                    \textbf{Real 20  +}             & Sim 20 & Sim 100 & Sim 200 & Sim 500 & Sim 1000 \\ \midrule
    \texttt{Real} + \texttt{DC} & 48\%              & 73\%                 & 85\%                & 88\%               & 95\%             \\
    \texttt{Real} + \texttt{DC w/ V2V}     & 70\%               & 80\%                 & 88\%                & 93\%               & 95\%             \\ 
    \bottomrule
    \toprule   
     \textbf{Sim 1000  +}             & Real 1 & Real 5 & Real 10 & Real 20 & Real 50 \\ \midrule
    \texttt{Real} + \texttt{DC}   & 8\%              & 40\%                & 73\%                & 95\%      & 95\%            \\
    \texttt{Real} + \texttt{DC w/ V2V}   & 13\%               & 53\%                & 83\%                & 95\%             & 95\%            \\ 
    \bottomrule
    \end{tabular}}
    \caption{\small{\textbf{Effects of improved visual realism with Vid2Vid models on policy performance in various numbers of simulation and real demonstrations.} We compare the policy co-trained on \texttt{Real} + \texttt{DC} for the \texttt{CupPnP} task where \texttt{DC} is augmented to be visually more realistic using the Vid2Vid model (\texttt{DC w/ V2V}) compared to using the OpenGL-based simulation renderings (\texttt{DC}).}}
    \label{tab:vid2vid}
    \vspace{-6pt}
\end{table}

\subsection{Training CloseDoor with More Demos}  
\label{app:closedoorextrademos}

Due to the large performance gap between \texttt{real} and all other co-trained policies for the \texttt{CloseDoor} task, we investigate whether this gap can be easily closed by training the \texttt{real} policy on more demonstrations. Specifically, we train the \texttt{real} policy on 100 demos instead of 50. We find that the policy still does not achieve 100\% success, only getting an 80\% success rate despite training with double demos.

\begin{figure}[t]
    \centering
    \includegraphics[width=\linewidth]{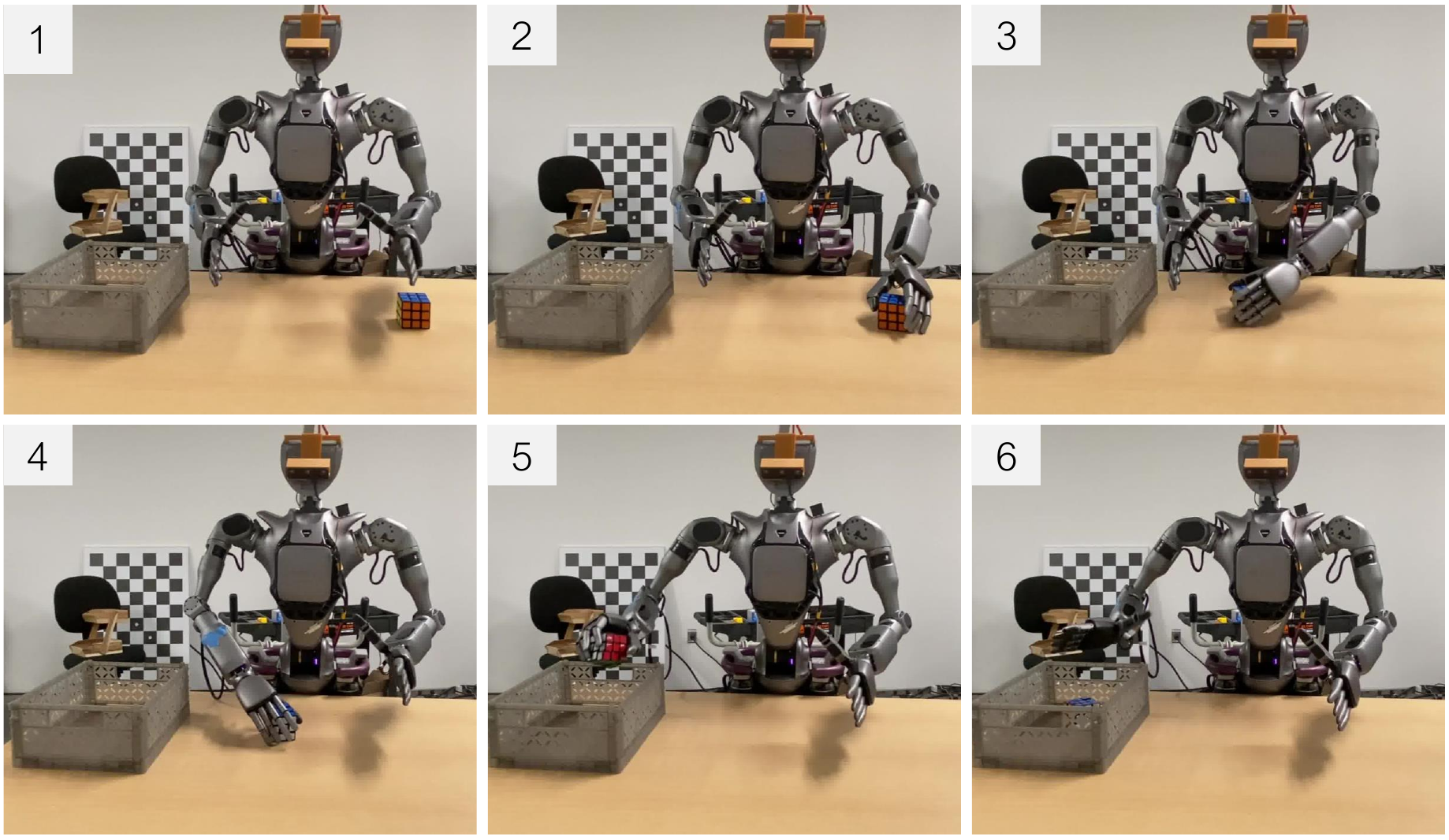}
    \caption{\textbf{Visualization of \texttt{BimanualPnP} task.} We show one rollout of our co-trained policy on \texttt{BimanualPnP} task. This task involves bimanual and long-horizon manipulation.}
    \label{fig:bimanual}
    \vspace{-10pt}
\end{figure}

\subsection{FAQs}
\label{app:faq}

\textbf{Is dynamic alignment investigated?}  
We have attempted to align dynamics by tuning physics parameters to reduce the gap between open-loop rollout results in simulation and the real world. However, our experiments on GR-1 \texttt{CupPnP} show no difference in success rate (95\%) with or without dynamic alignment, suggesting that such alignment is unnecessary for our tasks.  

\textbf{How misaligned is the default camera?}  
Since perfect camera alignment between simulation and the real world is infeasible, quantifying the exact camera pose error is challenging. Instead, we compute the delta pose between the default camera and the ``aligned'' camera. In Panda, the position and orientation deltas for the third-person cameras are 37 cm and $20^{\circ}$, respectively, while for the wrist camera, the deltas are 9 cm and $180^{\circ}$. In GR-1, the position delta is 36 cm, and the orientation delta is $60^{\circ}$. A visualization of the camera views is provided in Figure~\ref{fig:camera-view}.  

\textbf{How does co-training compare to domain randomization and domain adaptation?}  
We find that domain randomization and domain adaptation are complementary to co-training but not strictly necessary to reap the benefits of co-training with simulation. For instance, in the Panda Kitchen domain, we observe that adding generative texture randomization to the \texttt{digital cousin} data further improves performance, but even without it, co-training alone still leads to performance gains. Similarly, in the Humanoid Tabletop domain, enhancing visual realism via a Video2Video model can boost co-training results, but the gains are marginal in some cases. 

\textbf{Are there plans for bimanual tasks?}  
We introduce a GR-1 \texttt{BimanualPnP} task, where the humanoid must use its left hand to pick up a Rubik’s cube and place it at the center of the table, then use its right hand to pick up the cube and place it into a basket, as shown in Figure~\ref{fig:bimanual}.  
Our results show that with 50 real demonstrations, the policy achieves a 15\% success rate. When co-trained with 1,000 DC simulation demonstrations and 50 real demonstrations, the success rate improves to 50\%. In contrast, a policy trained only with 100 real demonstrations achieves a 30\% success rate.  
We also briefly tested co-training with our current task-agnostic prior datasets. As discussed in Section~\ref{app:priordata}, all our humanoid prior datasets consist of single-arm pick-and-place data. When co-trained with these datasets using a 99\% co-training ratio, the policy exhibits single-arm pick-and-place behavior instead of the expected bimanual behavior, resulting in a near-zero success rate. This suggests that for task-agnostic prior datasets to effectively support real-world manipulation, the behavior patterns in simulation and the real-world need to be consistent.